\documentclass[default,iicol]{sn-jnl}

\jyear{2021}%

\theoremstyle{thmstyleone}%
\theoremstyle{thmstyletwo}%

\theoremstyle{thmstylethree}%

\usepackage{subfigure}

\usepackage{xspace}

\newcommand{\eg}{\textit{e.g.}\xspace}
\newcommand{\etal}{\textit{et al.}\xspace}

\begin{document}

\title[Article Title]{Graph Fusion Network for Multi-Oriented Object Detection}


\author[1]{\fnm{Shi-Xue} \sur{Zhang}}

\author*[1]{\fnm{Xiaobin} \sur{Zhu}}
\author[1]{\fnm{Jie-Bo} \sur{Hou}}

\author[1,2]{\fnm{Xu-Cheng} \sur{Yin}}

\affil[1]{\orgdiv{School of Computer and Communication Engineering, USTB}}

\affil[2]{\orgdiv{Institute of Artificial Intelligence, USTB}}


\abstract{In object detection, non-maximum suppression (NMS) methods are extensively adopted to remove horizontal duplicates of detected dense boxes for generating final object instances. However, due to the degraded quality of dense detection boxes and not explicit exploration of the context information, existing NMS methods via simple intersection-over-union (IoU) metrics tend to underperform on multi-oriented and long-size objects detection. Distinguishing with general NMS methods via duplicate removal, we propose a novel graph fusion network, named GFNet, for multi-oriented object detection. Our GFNet is extensible and adaptively fuse dense detection boxes to detect more accurate and holistic multi-oriented object instances. Specifically, we first adopt a locality-aware clustering algorithm to group dense detection boxes into different clusters. We will construct an instance sub-graph for the detection boxes belonging to one cluster. Then, we propose a  graph-based fusion network via Graph Convolutional Network (GCN) to learn to reason and fuse the detection boxes for generating final instance boxes. Extensive experiments both on public available multi-oriented text datasets (including MSRA-TD500, ICDAR2015, ICDAR2017-MLT) and multi-oriented object datasets (DOTA) verify the effectiveness and robustness of our method against general NMS methods in multi-oriented object detection.
}


\keywords{Graph fusion network, Non-maximum suppression, Multi-oriented object detection}



\maketitle

\footnotetext{Published version Source: \url{https://link.springer.com/article/10.1007/s10489-022-03396-5}}

\section{Introduction}\label{sec1}
Object detection is a fundamental problem in computer vision and has been extensively studied. It can benefit various applications such as autonomous driving~\cite{drive2, drive1}, video or image indexing~\cite{Video_Google, Object_matching}, and scene parsing~\cite{Yin-Z, KangKY17}. 
Recent years have witnessed significant progress in object detection with the development of deep convolutional neural networks (CNNs). The state-of-the-art object detection methods ~\cite{DenseBox,R2CNN++,Zhang2022ArbitraryST,EAST, Zhang2022KernelPN} mostly detect an object  by generating dense detection boxes. Hence, duplicate removal is 
an important and inevitable post-processing procedure, which eliminates highly overlapped detection results and only retains the most accurate bounding box for each object.

As an essential object detection step, the duplicate removal strategy can significantly affect the final detection results. However, existing works often focus on generating accurate object proposals and corresponding class labels. Comparatively speaking, duplicate removal strategies have been seldom addressed. To remove the duplicate,  NMS~\cite{NMS} is still a popular and default solution, which iteratively selects proposals according to their prediction scores and suppresses overlapped proposals. Recently, Soft NMS~\cite{soft_nms} and learning-based NMS~\cite{learn_nms,Relation_Networks, RNN_nms, Softer-NMS} are proposed to improve NMS results. Instead of eliminating all lower scored surrounding bounding boxes, Soft-NMS~\cite{soft_nms} decays the detection scores of all other neighbors as a continuous function of their overlap with the higher scored bounding box. Jan~\etal~\cite{learn_nms} adopt the ConvNet to re-score all raw detections for searching suitable detection boxes. Relation network~\cite{Relation_Networks} regards duplicate removal as a two-class classification problem and performs the classification process by modeling the relation between object proposals with the same prediction class label. Softer-NMS~\cite{Softer-NMS} uses the learned variances of neighboring bounding boxes to weighted fuse the detections corresponding to the same object.

Both Soft-NMS~\cite{soft_nms} and Softer-NMS~\cite{Softer-NMS} can't directly predict the final detection by learning way, which predicts some auxiliary information to optimize NMS operation. And another problem is that the NMS algorithms mentioned above are excellent in horizontal object detection. Still, they can't be directly applied to multi-oriented object detection because they are designed according to the characteristics of horizontal object detection. In many scenes, multi-oriented objects are ubiquitous, such as objects in aerial images and scene texts. Consequently, horizontal bounding boxes don't provide accurate orientation and scale information, which poses a problem in real applications such as object change detection in aerial images and recognition of sequential characters for multi-oriented scene texts. At present, the popular way to deal with these dense rotated bounding boxes or quadrangle representations is still the greedy and hand-crafted non-maximum suppression, such as Locality-Aware NMS \cite{EAST} and Skew NMS~\cite{RRPN}.

\begin{figure}[ht]
	\subfigcapskip=0pt
	\centering
	\subfigure[Dense Boxes]{
		\begin{minipage}[t]{0.465\linewidth}
			\centering
			\includegraphics[width=3.55cm,height=3cm]{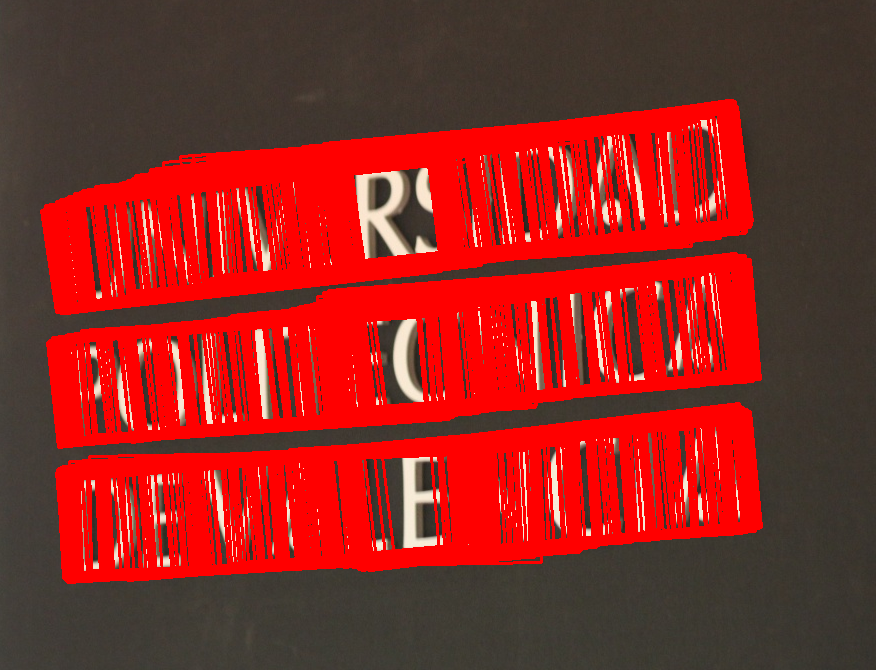}\\
		\end{minipage}%
	}%
	\subfigure[Skew NMS]{
		\begin{minipage}[t]{0.465\linewidth}
			\centering
			\includegraphics[width=3.6cm,height=3cm]{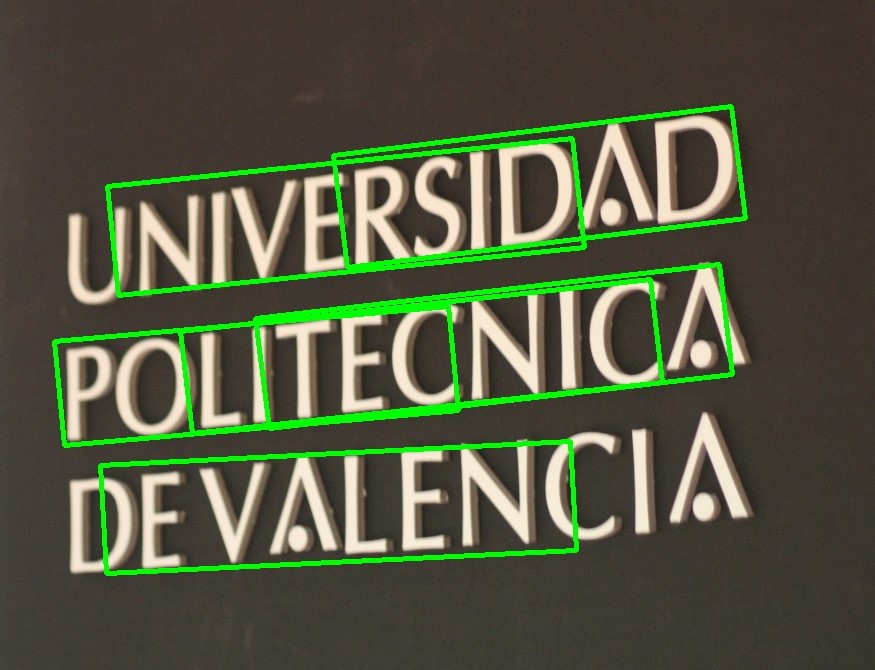}\\
		\end{minipage}
	}%
	\vspace{-3pt}
	\quad
	\subfigure[Locality-Aware NMS]{
		\begin{minipage}[t]{0.465\linewidth}
			\centering
			\includegraphics[width=3.65cm,height=3cm]{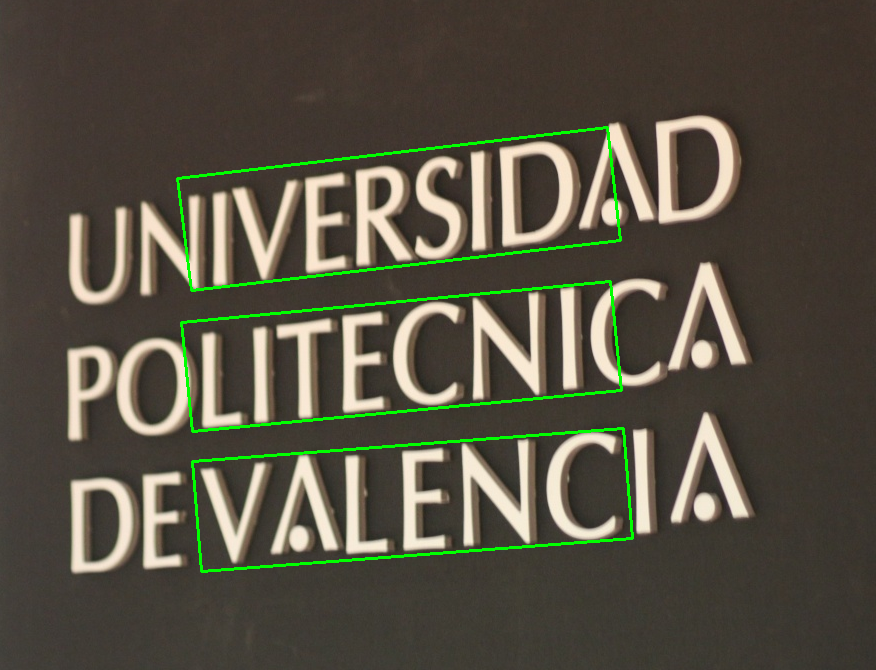}\\
		\end{minipage}
	}%
	\subfigure[Our GFNet]{
		\begin{minipage}[t]{0.465\linewidth}
			\centering
			\includegraphics[width=3.65cm,height=3cm]{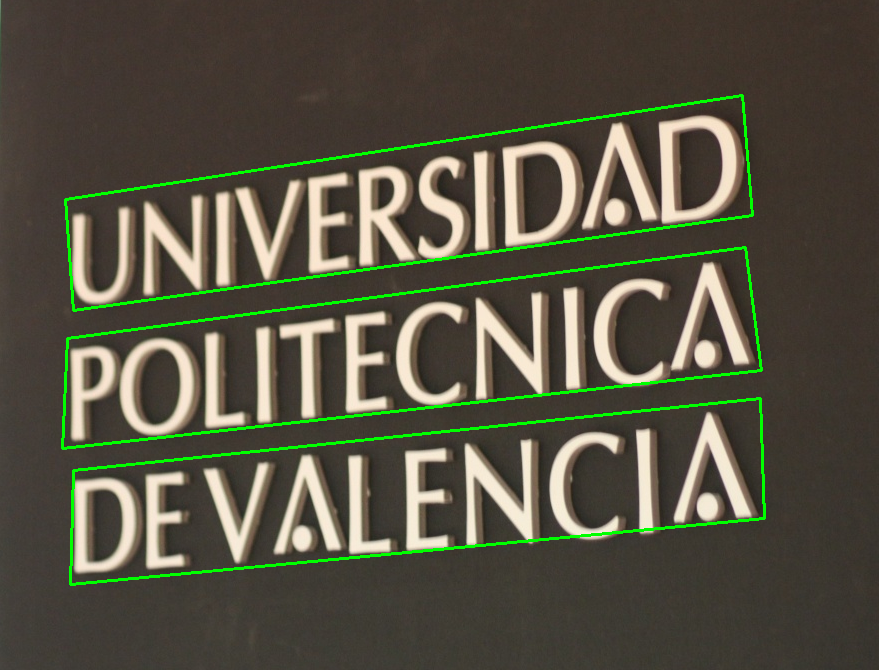}\\
		\end{minipage}
	}%
	\centering
	\caption{(a)~Dense Boxes generated by EAST~\citep{EAST}; (b)~Final predictions generated by Skew NMS~\citep{RRPN} on (a); (c)~Final predictions generated by Locality-aware NMS~\citep{EAST} on (a); (d)~Final predictions generated by our method on (a).}
	\label{fig:fg1}
\end{figure}

However, multi-oriented objects usually have arbitrary angles and variable aspect ratios, making it difficult for a detector to produce informative and accurate detection boxes. As shown in Fig.~\ref{fig:fg1}~(a), dense boxes generated by text detector EAST~\citep{EAST} are not holistic or well covered. It isn't easy to cover the whole text by only retaining the boxes with the highest score for each text. Hence, duplicate removal strategies (NMS methods) cannot generate promising object boxes, as shown in Fig.~\ref{fig:fg1}~(b). Locality-Aware NMS attempts to fuse these detection boxes by linear weighting via scores instead of simply removing duplicates with the confidence of text classification to improve the final detection boxes. Although the final result is still unsatisfactory, as shown in Fig.~\ref{fig:fg1}~(c), this strategy does bring some improvement compared with the duplicate removal strategy. Our observations and experiments believe that duplicate fusion strategy is more reasonable than duplicate removal strategy, especially on detecting a multi-oriented object or long object. Locality-Aware NMS and softer NMS have shown us the potential advantages of duplicate removal.

To address the above-mentioned problems, we propose a novel graph-based fusion network, named GFNet, for multi-oriented object detection. It can learn to adaptively fuse dense detection boxes for generating more accurate and holistic object instances on multi-oriented object detection. Specifically, we first adopt a locality-aware clustering algorithm to group dense boxes from the detector into different clusters based on the intersection over union (IoU) between detection boxes. A set of detection boxes can be seen as a graph where overlapping windows are represented as edges in a graph of detection boxes. To facilitate the network to learn fusion, we will construct an instance sub-graph for the detection boxes belonging to one cluster. Then, we propose a graph-based fusion network via Graph Convolutional Network (GCN) to learn to reason and fuse the detection boxes for generating final instance boxes. In summary, the main contributions of this paper are three-fold:

\begin{itemize}
	\item We propose a novel graph-based fusion network, GFNet, to adaptively fuse dense detection boxes to generate
	more accurate and holistic object instances for multi-oriented object detection. Our GFNet is extensible and can be easily generalized to other dense boxes oriented detection networks.
	
	\item We propose a novel instance sub-graph algorithm to explicitly explore and reason the relationships among dense detection boxes.
	
	\item Extensive experiments on multi-oriented object detection datasets, including scene text and aerial images, verify that our method is more effective and robust 
	against general NMS strategies in multi-oriented object detection.
\end{itemize}

The rest of this paper is organized as follows. We shortly review some related works on scene text detection in Section~\ref{Related_Work}.
The proposed method is then detailed in Section~\ref{Proposed_Method}, followed by extensive experimental results in Section~\ref{Experiments}. Finally, we conclude and give some perspectives on Section~\ref{Conclusion}.

\begin{figure*}[htbp]
	\begin{center}
		\includegraphics[width=0.99\linewidth]{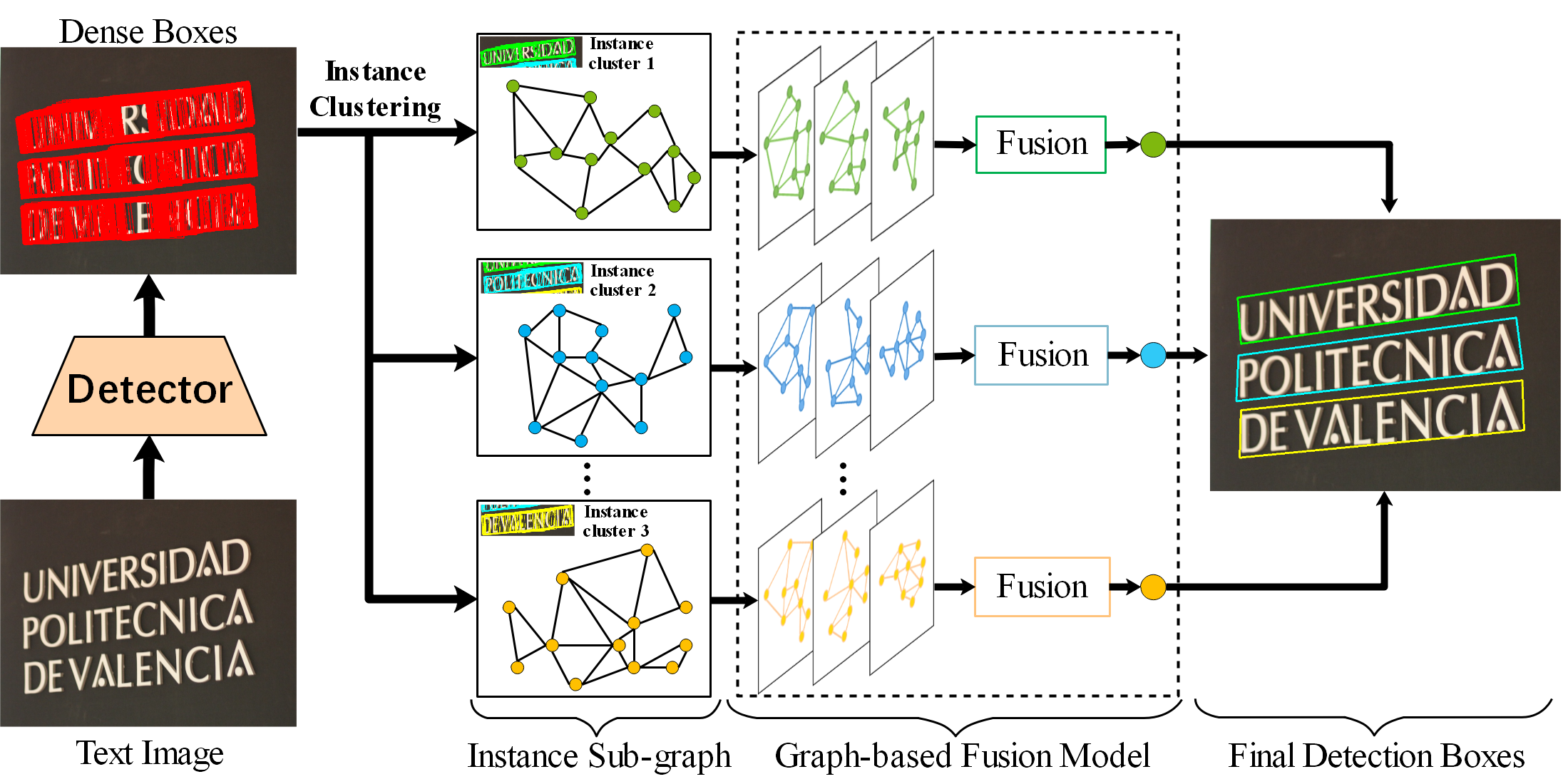}
		\caption{Framework of our method. We first adopt a detector to predict dense boxes. After locality-aware clustering, we construct an Instance Sub-graph for each object instance. Afterward, we apply a graph-based fusion model to learn to fuse the nodes in each sub-graph for generating final object instances}
		\label{fig:fig2}
	\end{center}
\end{figure*}

\section{Related Work} \label{Related_Work}
\small \subsection{\normalsize{Deep general object detection}}
General object detection methods~\cite{Context-aware,Instant-Teaching,PAQ} aim to detect an object in images with horizontal bounding boxes. In recent years, CNN-based object detection methods have achieved great success, including two-stage, single-stage, and keypoint-based methods. R-CNN and
its variances~\citep{Fast_RCNN, Faster-rcnn,R-FCN} are representative two-stage methods in which dense object proposals are first generated then be predicted and refined for final bounding boxes. YOLO and its variances~\cite{YOLO1,YOLO2} are representative single-stage methods, which directly predict bounding boxes without explicitly generating object proposals. As to keypoint-based methods (e.g., CenterNet~\cite{CenterNet} and TextBPN~\cite{TextBPN}), they attempt to explore informative key points of objects (such as corners and points) for grouping them into final bounding boxes.

\subsection{\normalsize{Multi-Oriented Object Detection}}
Multi-oriented object detection is challenging due to arbitrary orientations and variable aspect ratios. Multi-oriented text detection in scene images and multi-oriented object detection in the aerial image are two representative tasks in which the aspect ratio of text or object instance changes dramatically. 

In multi-oriented scene text detection, regression-based methods are popular, including anchor-based methods~\citep{textboxes++,RRPN} and anchor-free methods~\cite{EAST,DDR,MOST,HAM}. They usually directly predict entire texts using a rotated bounding box or quadrangle. RRPN~\citep{RRPN} employ rotated RPN in the framework of Faster R-CNN~\citep{Faster-rcnn} to generate rotated proposals and further perform rotated bounding box regression. EAST~\citep{EAST} and DDR~\citep{DDR} perform rotated bounding box regression or vertex regression at each location. MOST~\cite{MOST} puts forward a set of strategies to improve the quality of text localization for long text significantly.

In aerial object detection, extensive
studies conduct their experiments on the popular dataset DOTA~\citep{DOTA}. Ding \etal~\citep{RoI_Transformer} proposed an RoI transformer that transforms horizontal proposals into rotated ones for performing the rotated bounding box regression.  R$ ^{2} $CNN++~\citep{R2CNN++} fuses multi-layer features with effective anchor sampling to improve the sensitivity of the model to small objects. Pan \etal~\citet{DRN} presented a dynamic refinement network that consists of two novel components, i.e., a feature selection module (FSM) and a dynamic refinement head (DRH), to solve the problem of multi orientations and generalization of the model.

\subsection{\normalsize{Non-Maximum Suppression Methods}}
\textbf{Duplicate Removal of Horizontal Boxes.} Duplicate removal is an indispensable step in many object detection frameworks. To remove duplicates,  NMS~\citep{NMS} will iteratively select proposals according to their classification confidence scores and suppress overlapped proposals by IoU strategy.  Traditional hard NMS traditional strategies do not consider any context information and inter-relations between bounding boxes. Hence they often abandon useful candidates. Soft-NMS~\citep{soft_nms} decays the detection scores of all other neighbors as a continuous function of their overlap with the higher scored bounding box. The limitation is that many redundant proposals may still be
kept in the final prediction. Recently, learning-based NMS methods~\citep{learn_nms,Relation_Networks,RNN_nms} have been proposed and achieved promising performance. In~\citet{learn_nms}, a non-maximum suppression ConvNet is proposed to re-score all raw detections for search suitable detection boxes. Relation network~\citep{Relation_Networks} regards duplicate removal as a two-class classification problem, which models the relation between object proposals with the same class label and uses learned implicit relations to divide these proposals into two categories. Qi, \etal~\citet{RNN_nms} 
adopted recurrent neural networks (RNN) to model structure information underlying proposal candidates for duplicate removal.

\textbf{Duplicate Removal of Oriented Boxes.} Although the state-of-the-art NMS algorithms achieve promising performances in horizontal object detection, their performances may greatly drop in the task of multi-oriented detection. To remove duplicates in multi-oriented object detection, RRPN~\citep{RRPN} adopt Skew NMS, which design implementation for the skew IoU computation with consideration of the triangulation to dense rotated bounding boxes or quadrangle representations. R$ ^{2} $CNN++~\citep{R2CNN++} adopt rotation non-maximum suppression (R-NMS) as a post-processing operation based
on skew IoU computation. To apply NMS to polygon duplicate removal, Liu~\etal~\citep{CTW1500} proposed a polygonal non-maximum suppression (Polygon NMS) which improves the traditional NMS by computing the overlapping area between polygons instead of horizontal quadrilateral bounding boxes. 

\subsection{\normalsize{Duplicate Fusion Methods}} 
In the task of multi-oriented object detection, dense boxes generated by the detector are always inaccurate. Consequently, a duplicate removal strategy can easily abandon useful boxes for final prediction. Under the assumption that the geometry relations of nearby pixels tend to be highly correlated, Locality-Aware NMS~\citep{EAST} merge
the dense detection boxes by linear weighting strategy according to their classification confidences. Softer-NMS~\cite{Softer-NMS} uses the learned variances of neighboring bounding boxes to weighted fuse the detections corresponding to the same object to replace the traditional NMS.  Although the duplicate fusion strategies are useful for multi-oriented object detection, there are still many problems that remain to be solved, and few works have been reported. More existing methods focus more on improving the quality of dense detection boxes directly generated by the network. Therefore, the research of duplicate fusion is left far behind.

\subsection{\normalsize{Graph Convolutional Network}}
In the past few years, the research~\citep{GCN, GCN_2, GAT, GCN_1, CVPR19_Linkage} of the convolutional neural network for graph-structured data has made remarkable achievements. According to the definition of convolution on graph data, GCNs can be categorized into spectral methods and spatial methods. Spectral-based GCNs~\citep{GCN, GCN_2, GAT} generalize convolution based on Graph Fourier Transform, while spatial-based GCNs~\citep{GCN_1, CVPR19_Linkage} directly perform manually-defined convolution on graph nodes and their neighbors. In our work, a set of detection boxes can be seen as a graph where overlapping windows are represented as edges in a graph of detection boxes. Hence, we propose a spatial-based GCN model to fuse the detection boxes based on the instance sub-graph.

\begin{figure}[htp]
	\begin{center}
		\includegraphics[width=0.98\linewidth]{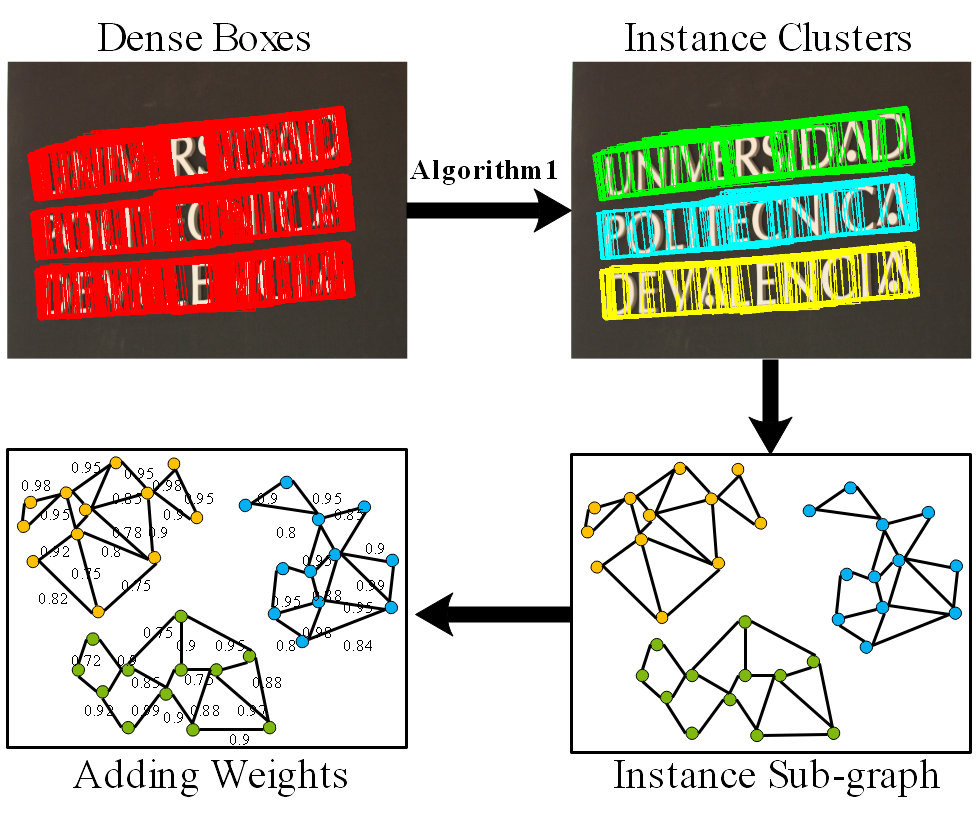}
		\caption{Diagram of instance sub-graph construction. Dense boxes will be predicted by a detector and then be clustered according to Algorithm.~\ref{algo:algorithm_1}. Based on the clustering results, we will construct an Instance Sub-graph via dense boxes as nodes. The IoU between two nodes determines the edge weight in the sub-graph.}
		\label{fig:fig3}
	\end{center}%
\end{figure}

\section{Our Method}\label{Proposed_Method}
\subsection{\normalsize{Overview}}
The framework of our method is depicted in Fig.~\ref{fig:fig2}. Firstly, we adopt a regression-based detector to generate dense boxes. Then, we propose a locality-aware clustering algorithm to group dense boxes into different clusters via the IoU metric. Regarding the dense boxes of one object instance as nodes, we can construct an instance sub-graph for each object instance according to their geometry relationships. Finally, we adopt a graph-based network to explore the spatial context information between dense boxes to perform duplicate fusion for further optimizing final detection results.


\begin{algorithm}
	\caption{Locality-aware Clustering}\label{algo:algorithm_1}
	\begin{algorithmic}[1]
		\Require $ B=\{b_0,...b_i,...,b_n\} $,  $ th_{iou} $;
		\Statex $ B $ is a set of dense boxes; $ b_i $ is a 9-dimension scalar; $ th_{iou} $ is the threshold of IoU;
		\State $ p \gets Null $,
		$ \mathcal{S} \gets \emptyset $, $ s \gets \emptyset$;
		
		\While{$ B \neq \emptyset$}
		\State $ p \gets B.pop() $;
		\State $  s \gets  s \cup \{  p \} $;
		\State $ \mathcal{B} \gets \emptyset $ ;
		\For{$ g \in B $}
		
		\If{$ IoU(g, p) \geq  th_{iou}$}
		\State $p = Merge(g,p) $;
		\State $  s \gets  s \cup \{ g \} $;
		\Else
		\State $  \mathcal{B} \gets \mathcal{B} \cup \{ g \} $;
		\EndIf
		
		\EndFor
		\State  $ B \gets \mathcal{B}$;
		\If{$  s \neq \emptyset $}
		\State $  \mathcal{S} \gets \mathcal{S} \cup s $;	$ s \gets \emptyset$;
		\EndIf
		\EndWhile
		\If{$  S  =  \emptyset $}
		\State $  S  \gets  Null $;
		\EndIf \\
		\Return $\mathcal{S}$	
	\end{algorithmic}
	\vspace{1.0em}
\end{algorithm}

\subsection{\normalsize{Instance Sub-graph Construction}} 
The diagram of instance sub-graph construction is depicted in Fig.~\ref{fig:fig3}. We first group the detected dense boxes into clusters belonging to different possible object instances for constructing instance sub-graph. Then, we will uniformly sample boxes for controlling the complexity of our graph network. Regarding dense boxes as nodes, we will construct an Instance Sub-graph for each object instance, in which the IoU between two nodes determines the edge weight.

\medskip
After getting the dense detection boxes, we need to find all the detection boxes belonging to one object instance for constructing construct instance sub-graphs. Inspired by~\citep{EAST}, we propose a locality-aware algorithm to group dense detection boxes, which is detailed in Algorithm~\ref{algo:algorithm_1}. The scalar $ b_i $ contains two parts of information $ (v, s) $, in which $ v=(x_1, y_1, x_2,y_2,x_3,y_3,x_4,y_4)$ denote  four vertex coordinate information of a detection box, and $ s $ denotes a prediction score of a detection box. To cluster these detection boxes, we merge the geometries row by row under the assumption that geometry relationships between nearby pixels tend to be highly correlated. And while merging geometries in the same row, we will iteratively merge the geometry currently encountered with the last merged one, which makes our cluster algorithm run in less than $O(n^2)$. To merge the geometries of detection boxes, we use the scores of two given quadrangles to weight-average the coordinates of the merged quadrangle. Supposing $p^{*} = Merge(g, p)$, the specific calculation process of $Merge$ is as follows:
\begin{gather}
\medskip
s_{p^{*}}= s_{g}+ s_{p},\\
v_{p^{*}}= (s_{g}*v_{g} + s_{p}*v_{p})/s_{p^{*}},
\medskip
\end{gather}
where $s_{p^{*}}$ is the score of new box $  p^{*} $, and $v_{p^{*}}$ is one of the coordinates of new box $p^{*}$; $g$ is one of the detection boxes, and $p^{*}$ is the merged box generated by $Merge$. Here are some visual examples in Fig.~\ref{fig:fig3}. The dense detection boxes are shown with red quadrilateral, and the clustering results are shown with green, blue, and yellow quadrilateral in instance cluster.

\medskip

Through object instance clustering, we can group dense boxes into different clusters (different object instances). In general, each object instance will have a large number of detection boxes. And the number of detection boxes may greatly be varied according to the size of an object instance. Unfortunately, the computation cost of the graph convolution network will be dramatically increased along with the increase of nodes. To reduce the computation cost and improve the efficiency of our graph network, we fix the number of detection boxes as the nodes for each instance sub-graph.
Specifically, we first rank all the detection boxes of an object instance according to their scores and then uniformly sample $ N $ detection boxes. In our experiments, $ N $ is set to 128 for EAST detector and 64 for R$ ^{2}$CNN++ detector.

\medskip
The last step is to build connections among the nodes. For instance sub-graph $ G(V, E) $, we first compute the intersection over union (IoU) between nodes, and then we can get a matrix ($ W $) of $ N \times N $. In matrix $ W $, each number represents IoU between two corresponding nodes in the sub-graph. To obtain the adjacent matrix ($ A $) of the instance sub-graph, we adopt a pre-defined threshold ( $ th_{iou}=0.7 $ in our experiments) to binary the matrix $ W $. It is worth noting that the diagonal element of matrix $ W $ is always 1 because the node and its own IoU are 1 permanently.
Hence, the diagonal elements of adjacency matrix $ \tilde{A} $ $ (\tilde{A}=A+I_N) $ are all 1 too.

\medskip
For each instance sub-graph node (detection box), we use the normalized geometric attributes of the detection box as the feature of the node. It is easy for us to get four vertex coordinates of each detection box directly from the predictions of the detector. Then we use the known image size to normalize this set of geometric attributes $ (u=(x_c,y_c, x_1, y_1, x_2,y_2, x_3,y_3,x_4,y_4)) $. The specific normalization operation is as follows:

\begin{gather}
\medskip
u_{x_c} = x_c/w; u_{y_c} = y_c/h,\\
u_{x_i} = (x_i-x_c)/w; u_{y_i} = (y_i-y_c)/h;
\medskip
\end{gather}
where $ w $ and $ h $ respectively denote the width and height of an image; $ x_c $ and $ y_c $ respectively denote the coordinates of the center point of the corresponding detection box;  $ x_i $ and $ y_i $ respectively denote the coordinates of the detection box points. And, $  u_{x_c} $,$ u_{y_c} $, $ u_{x_c} $ and $ u_{y_c} $ are the corresponding normalized values.

\subsection{\normalsize{Graph-based Fusion  Model}}\label{gcn_model}

To fully utilize and explore the context  information between nodes in sub-graph, we propose a graph-based fusion model based on graph convolution network~\citep{GCN} to learn to accurately fuse the detection boxes for each object instance, as shown in Fig.~\ref{fig:fig2}. Our graph-based fusion model mainly consists three graph convolution layers and node information fusion layer. In our work, we introduce residual structure and weighted aggregation matrix into our graph convolution network, which can be formulated as
\medskip
\begin{equation}
\mathbf{ Y^{(l)} = \sigma((X^{(l)} \oplus G X^{(l)})W^{(l)}}),
\medskip
\end{equation}
where $ \mathbf{{X}^{(l)}} \in \Re^{N \times d_{i}} , \mathbf{ { Y}^{(l)}}\in \Re^{N \times d_{o}}  $, $ d_i/d_o $ is the dimension of input / output node features and $ N  $ is the number of nodes; $ \mathbf{G} $ is a aggregation matrix of
size $ N \times N $; $ \oplus $ represents matrix concatenation; $  \mathbf{ W^{(l)}} $ is a layer-specific trainable weight matrix;
$ \sigma(\cdot) $ denotes a non-linear activation function. In our method, the  graph convolution layers activated by the ReLU function. The graph convolution operation can be broken down into two steps. In the first step, by left multiplying $ \mathbf{ X^{(l)}} $ by $ \mathbf{ G }$, the underlying information of node's neighbors is aggregated. Then, the input node features $ \mathbf{ X^{(l)} }$ are concatenated with the aggregated information $ \mathbf{ { GX^{(l)}} }$ along the feature dimension. In the second step, the concatenated features are transformed by a set of linear filters, whose parameter $ \mathbf{ W^{(l)} }$ is to be learned. 

We multiply the symmetric normalized Laplacian ($ \bold{L} $)  and a weight matrix ($ \bold{W} $) as the weighted aggregation matrix $ \bold G $. The symmetric normalized Laplacian ($ \bold{L} $) is obtained by adjacency matrix ($ \bold A $) transformation. For an instance sub-graph, we compute the intersection over union (IoU) between nodes, and then we can get a matrix ($ \bold W $) which is used as a weighted matrix. The weighted aggregation performs weighted average pooling among neighbors. The specific computation of aggregation matrix $ \bold G $ is as follows:
\medskip
\begin{gather}
\bold{L} =\bold{\tilde{D}}^{-1/2}\bold{\tilde{A}}\bold{\tilde{D}}^{-1/2},\\
\bold G = \bold{L}*\bold{W},
\medskip
\end{gather}
where $ \tilde{A} = A + I_N $ is an adjacency matrix of the instance sub-graph with added self-connections; $ I_N $ is the identity matrix and $ \bold{\tilde{D}} $ is a diagonal matrix with $ \tilde{D}_{ii} = \sum_{j}\tilde{A}_{ij}$.

To fuse the features extracted by the graph convolution layer to generate a new detection box for each object instance, we apply $ 1 \times 1 $ convolutions on the feature matrix $ \bold Y_g $ extracted from graph convolution layer along the dimensions of nodes.
\medskip
\begin{equation}
\bold Y=Conv_{1\times 1}(\bold Y_g),
\medskip
\end{equation}
where $ \bold Y_g $ is a feature matrix of
size $ B \times N \times 10 $, and $ \bold Y $ is the final predictions of size $ N \times 10 $. In this way, we can effectively predict a new node based on the instance sub-graph. In other words, we successfully fuse the multiple detection boxes into a new detection box.

\subsection{\normalsize{Optimization}}
The proposed graph-based fusion network infers a unfixed-size set of $ M $ predictions, and different image has a different number of the predictions $ M $. For computing the losses, we need to produce an optimal bipartite matching between predictions and ground-truths. Let $ u_i $ denote the ground truth set of detection boxes, and $ \hat{u}=\{\hat{u}_i\}_{i=1}^{M} $ denote the
set of $ M $ predictions. To find a bipartite matching between these two sets, we search for a permutation of M
elements $ \varepsilon \in E_{M} $ with the lowest cost:
\medskip
\begin{gather}
\hat{\varepsilon} = \mathop{\arg\min}\limits_{\varepsilon \in E_{M}} \sum_{i=1}^{M}\mathcal{L}_{box}(u_{i}, \hat{u}_{\varepsilon (i)})
\end{gather}
\medskip
where $ \mathcal{L}_{box}(u_{i}, \hat{u}_{\varepsilon (i)}) $ is the loss between ground truth $ u_i $ and a prediction with index $ \varepsilon(i) $. 

The bounding box points regression losses are performed by an Euclidean loss. The scalar $ u=(u_{x_c},u_{y_c},\\u_{x1}, u_{y1}, u_{x2},...,u_{x4},u_{y4}) $ is a label tuple of coordinates of detection box points, and the scalar $ \hat{u}=(\hat{u}_{x_c}, \hat{u}_{y_c}, \hat{u}_{x_1},\\ \hat{u}_{y_1},...,\hat{u}{x_4},\hat{u}_{y_4}) $ is the predicted tuple for the detection box point. To make the points learned
suitable for object instance of different scales, the learning targets should also be processed to make them scale invariant. The parameters $ (u_{x_c}, u_{y_c}) $ and $ (u_{x_i}, u_{y_i}) $ are processed as following:
\medskip
\begin{gather}
u_{x_c} = x_c/w; u_{y_c}= y_c/h,\\
u_{x_i} = (x_i-x_c)/w; u_{y_i} = (y_i-y_c)/h,
\end{gather}
\medskip
where $  x_i $ and $ y_i $
denote the coordinates of the detection box
points, $ x_c  $ and $ y_c $ denote the coordinates of the center point
of the corresponding detection box, $ w $ and $ h$ denote the width and height of this image.
We adopt the smooth L1 loss~\citep{Faster-rcnn} as the loss function in our method:
\medskip
\begin{gather}
\mathcal{L}_{box}(u, \hat{u}) = \sum_{i
	=1}^{M}Smooth_{L1}(u,\hat{u}_{\hat{\varepsilon} (i)}),\\
\medskip
Smooth_{L1}(x) =
\begin{cases}
0.5x^2,\; if |x|<th,\\
|x| - 0.5,\;otherwise,
\end{cases}
\medskip
\end{gather}
where $ th $ is set to 0.33 in our experiments.

\section{Experiments}\label{Experiments}
In this section, we conduct experiments on public multi-oriented scene text detection datasets based on the text detector EAST \citep{EAST} and multi-oriented object detection datasets based on the method R$ ^{2} $CNN++ \citep{R2CNN++}. We show comparative results and perform sensitivity analysis to show the robustness of our GFNet Compared to traditional NMS. Some implementation details are depicted in Section~\ref{Implement-Details}, followed by comparison experiments for text detection in scene images in Section~\ref{text_detect_exp} and comparison experiments for object detection in aerial images in Section~\ref{object_detect_exp}. The weakness analysis is given in Section~\ref{Weakness}.

\subsection{\normalsize{Datasets}}
\textbf{ICDAR2015}~\citep{IC15}:
It is a popular dataset released for the Robust Reading Competition, which consists of 1,000 training images and 500 testing images, all annotated by word-level bounding quadrilaterals. Google Glasses take ICDAR2015 in an incidental manner. Therefore, texts in these images are of various scales, orientations, contrast, blurring, and viewpoint, making it challenging for detection.

\medskip
\textbf{MSRA-TD500}~\citep{MSRATD500}: 
This dataset is often used in the task of multi-lingual arbitrary orientation long text detection. It consists of 500 training images and 200 testing images with multi-lingual long texts annotated with line-level. 

\medskip
\textbf{ICDAR2017-MLT}~\citep{MLT}: This dataset is a large-scale multi-lingual text dataset focusing on multi-oriented and multi-lingual text images (including 7,200 training images (84,868 cropped words), 2,000 validation images, and 9,000 testing images (97,619 cropped words)). It contains
9 languages (\eg, Chinese, Japanese, Korean, English, French, Arabic, Italian, German, and Indian). Similar to ICDAR2015, the text regions in ICDAR2017-MLT are also annotated by the 4 vertices of quadrilaterals. We use both our training set and the validation set in the training period.

\medskip

\textbf{DOTA}~\citep{DOTA}: This dataset is often used in the task of aerial image detection. It contains 2,806 aerial images captured by different sensors and platforms. The image resolution ranges from $ 800 \times 800 $ to $ 4000 \times 4000 $ pixels. Hence objects in images will exhibit a wide variety of scales, orientations, and shapes. These images contain 15 common object categories. The fully annotated DOTA benchmark has 188,282 instances which are all labeled by an arbitrary quadrilateral. There are two detection tasks for DOTA: horizontal bounding boxes (HBB) and oriented bounding boxes (OBB). Generally, half of the images are randomly selected for training, 1/6 of the images in the validation dataset will be selected for validation, and 1/3  of the images in the testing dataset will be selected for testing. 

%

\subsection{\normalsize{Implementation Details}}\label{Implement-Details}

Our method is extensible and can be well generalized to other models to deal with dense boxes and optimize the final detection boxes actively. To verify the effectiveness of our method, we conduct experiments on two different scenarios: muti-oriented text detection in scene images and muti-oriented object detection in the aerial image. For text detection, we adopt the popular scene text detection method EAST~\citep{EAST} as our detector. For multi-oriented object detection in aerial image, we adopt the popular multi-oriented object detection method R$ ^{2} $CNN++~\citep{R2CNN++} as our detector.

\textbf{Text Detector.} For multi-oriented scene text detection, we use the Python implementation of EAST \citep{EAST}\footnote{https://github.com/argman/EAST} For generating text detection boxes and made some improvements to get a better baseline. Specifically, we replace the backbone with more sophisticated ResNet50~\citep{ResNet} and adopt the online hard negative mining (OHEM)~\citep{OHEM} to overcome the unbalance number of positives and negatives. For training EAST, ResNet-50 will be pre-trained on ImageNet, and Adam \citep{ADAM} optimizer is adopted with a learning rate initialized 0.0001, which will decay with 0.94 factor after every 10k iterations. Also, basic data augmentation techniques such as resize, rotations, and crops are applied.

\textbf{Object Detector.} For multi-oriented object detection in aerial images~\citep{DOTA}, we use the Python implementation of R$ ^{2} $CNN++\footnote{\label{R2CNN++}https://github.com/DetectionTeamUCAS/R2CNN-Plus-Plus\_Tensorflow}~\citep{R2CNN++} for generating detection boxes. We directly downloaded the model with the default setting from their official Github, but our testing performance is slightly lower than reported.

\textbf{Training GFNet.} We train our GFNet with Adam optimizer with 100k iterations. The initial learning rate is set to $ 10^{-4} $, which will be decayed with 0.94 factor after every 10k iterations. We use the training set and the validation set to train the detector and then use the dense boxes generated by the detector without the typical non-maximum suppression step as the train data for our GFNet. Data augmentation techniques, including rotation, scaling, and cropping of the input image, are adopted for training network. Besides, during detecting object proposals, we adopt a low score threshold for keeping more candidates for GFNet.

Experiments are performed on a single GPU (GTX 1080Ti), Intel Xeon Silver 4108 CPU @ 1.80GHz, and Tensorflow 1.4.0. For a fair comparison, all compared NMS algorithms are implemented with open source code in GitHub.

\medskip

\begin{table}[htbp]
	\begin{center}
		\renewcommand{\arraystretch}{1.0}
		\caption{Experimental results on ICDAR2015. The symbol $^*$  means that the model is re-implemented. The “R”, “P” and “F” represent the Recall, Precision and F-measure, respectively. “P-NMS”, “S-NMS” and “L-NMS” represent Polygon NMS, Skew NMS and Locality-Aware NMS, respectively.}
		\label{table:ICDAR15}
		\resizebox{0.485\textwidth}{!}{
			\begin{tabular}{|l||c|c|c|}
				\hline
				\textbf{Methods}& \textbf{R}& \textbf{P} & \textbf{F}\\
				\hline
				SegLink \cite{SegLink} &76.8 &73.1 &75.0\\
				MCN \cite{MCN} &72 &80 &76\\
				RRPN \citep{RRPN} &77 &84 &80\\
				EAST \citep{EAST} &78.3 &83.3 &80.7\\
				RRD \citep{RRD} &79 &85.6 & 82.2\\
				TextField \citep{TextField} &80.05 &84.3 &82.4\\
				TextSnake \citep{TextSnake} &84.9 &80.4 &82.6\\
				Textboxes++\citep{textboxes++} &78.5 &87.8 &82.9\\
				PixelLink \citep{PixelLink} &82.0 &85.5 &83.7\\
				FTSN \citep{FTSN}  &80.0 &88.6 &84.1\\
				FOTS \citep{FOTS}&82.04 &88.84 &85.31\\
				PSENet-1s\citep{CVPR19_PSENet} &84.5 &86.92 & 85.69\\
				DRRG~\citep{DRRG}&84.69 &88.53 &86.56\\
				LSE \citep{CVPR19_LSA} &85.0 &88.3 &86.6\\
				ATRR \cite{CVPR19_ATRR} &83.3 &90.4 &86.8\\
				CRAFT \citep{CRAFT} &84.3 &89.8 &86.9\\
				ContourNet~\citep{ContourNet}&86.1 &87.6 &86.9\\
				LOMO \citep{CVPR19_LOMO} &83.5 &\textbf{91.3} &87.2\\
				\hline
				\hline
				\textbf{EAST$ ^{*} $}(P-NMS)&85.22&87.62&86.40\\
				\textbf{EAST$ ^{*} $}(S- NMS)&78.77&82.96&80.81\\
				\textbf{EAST$ ^{*} $}(L- NMS)&84.93&87.98&86.43 \\
				\textbf{EAST$ ^{*} $(GFNet)}&\textbf{85.14}&89.47& \textbf{87.25}\\
				\hline
		\end{tabular}}
	\end{center}%
\end{table}

\begin{table}[htbp]
	\begin{center}
		\renewcommand{\arraystretch}{1.0}
		\caption{Experimental results on MSRA-TD500. The symbol $^*$  means that the model is re-implemented. The “R”, “P” and “F”  represent the Recall, Precision and F-measure, respectively. “P-NMS”, “S-NMS” and “L-NMS” represent the Polygon NMS, Skew NMS and Locality-Aware NMS, respectively.}
		\label{table:TD500}
		\resizebox{0.485\textwidth}{!}{
			\begin{tabular}{|l||c|c|c|c|}
				\hline
				\textbf{Methods}& \textbf{R}& \textbf{P} & \textbf{F}\\
				\hline
				EAST~\citep{EAST} &67.43 &87.28 &76.08\\ 
				SegLink \citep{SegLink}  &70.0 &86.0 &77.0\\
				PixelLink \citep{PixelLink} &73.2 &83.0 & 77.8\\
				TextSnake \citep{TextSnake} &73.9 &83.2 &78.3\\
				Border \citep{ASTD} &77.4 &83.0 &80.1\\
				ITN~\citep{ITN} &72.3 &\textbf{90.3} &80.3\\
				Lyu et al. \citep{corner} &76.2 &87.6 &81.5\\
				TextField \citep{TextField}&75.9 &87.4 & 81.3\\
				MSR~\citep{MSR} &76.7 &87.4 &81.7\\
				FTSN~\citep{FTSN} &77.1 &87.6 &82.0\\
				LSE~\citep{CVPR19_LSA} &81.7& 84.2 &82.9\\
				CRAFT~\citep{CRAFT} &78.2 & 88.2 &82.9\\
				MCN \citep{MCN} &79 &88 &83\\
				IncepText~ \citep{IncepText} &79.0 &87.5 &83.0\\
				ATRR\citep{CVPR19_ATRR} &82.1 &85.2 & 83.6\\
				MCN-4\citep{MCN_ijcv} &80.1 &87.4 & 83.6\\
				PAN \citep{PSENet_v2}&\textbf{83.8} &84.4  &84.1\\
				\hline
				\hline
				\textbf{EAST$ ^{*} $}(P-NMS)&78.09&80.28&79.17\\
				\textbf{EAST$ ^{*} $}(S-NMS)&74.79&77.47&76.11\\
				\textbf{EAST$ ^{*} $}(L- NMS)&79.69&81.78&80.21\\
				\textbf{EAST$ ^{*} $(GFNet)}&81.75&86.63&\textbf{84.12}\\
				\hline
		\end{tabular}}
	\end{center}%
\end{table}

\subsection{\normalsize{Text Detection in Scene Images}}\label{text_detect_exp}
To verify the effectiveness of our method, we conduct experiments on three publicly available scene text detection datasets (\eg, MSRA-TD500, ICDAR2015, ICDAR2017-MLT) on EAST~\citep{EAST} detector. Specifically, we adopt the traditional NMS algorithms (including Polygon NMS, Skew NMS, and Locality-Aware NMS) and our GFNet to deal with the dense boxes detected by text detector EAST, respectively. We comprehensively investigate the performance and efficiency of different NMS algorithms. In addition, we also compare the performance of the state-of-the-art methods in terms of reported performance.

\textbf{ICDAR2015.} Images in this dataset are of low resolution and contain many small multi-oriented text instances. We first resize all mage to the resolution of $ 1920 \times 1024 $ for detection during testing. The detailed results are listed in Tab.~\ref{table:ICDAR15}. Compared with Polygon NMS, the performance of Skew NMS degrades 5.59\% on F-measure. Because ICDAR2015 is mainly composed of small texts, the quality of detection boxes generated by the detector is relatively accurate. So, Locality-Aware NMS based on linear weighted fusion doesn't show great advantages compared with Polygon NMS. Comparatively, our GFNet can learn to fuse the detection boxes, which is more robust and adaptive than Locality-Aware NMS. Therefore, our GFNet outperforms Locality-Aware NMS by 0.82\% in terms of F-measure. In addition, equipped with the proposed GFNet, \textbf{EAST*}  can achieve a promising  F-measure score 87.25\%, surpassing the state-of-the-art methods, such as LOMO~\citep{CVPR19_LOMO}, ATRR~\citep{CVPR19_ATRR}, CRAFT ~\citep{CRAFT} and ContourNet~\citep{ContourNet}.

\medskip

\begin{figure*}[htbp]
	\subfigcapskip=5pt
	\centering
	\subfigure[Dense Boxes]{
		\begin{minipage}[t]{0.2\linewidth}
			\centering
			\includegraphics[width=1.02\linewidth]{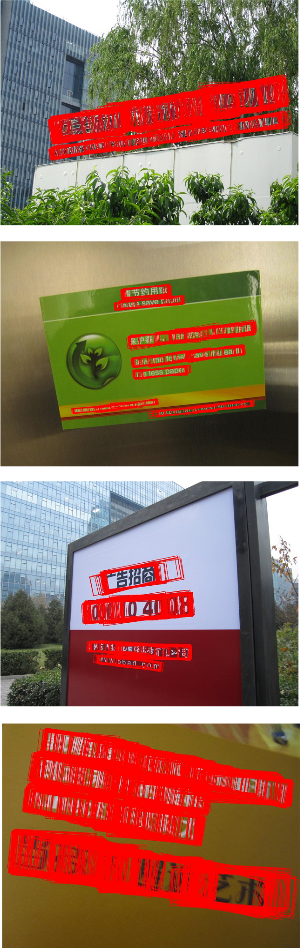}\\
		\end{minipage}%
	}%
	\subfigure[Polygon NMS]{
		\begin{minipage}[t]{0.2\linewidth}
			\centering
			\includegraphics[width=1.02\linewidth]{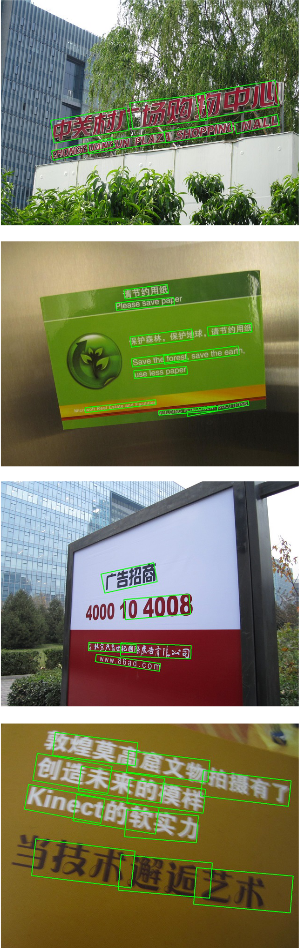}\\
			
		\end{minipage}%
	}%
	\subfigure[Skew NMS]{
		\begin{minipage}[t]{0.2\linewidth}
			\centering
			\includegraphics[width=1.02\linewidth]{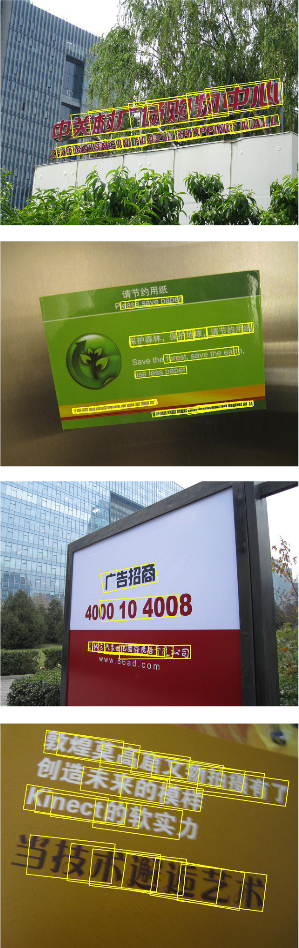}\\
		\end{minipage}%
	}%
	\subfigure[Locality-Aware NMS]{
		\begin{minipage}[t]{0.2\linewidth}
			\centering
			\includegraphics[width=1.02\linewidth]{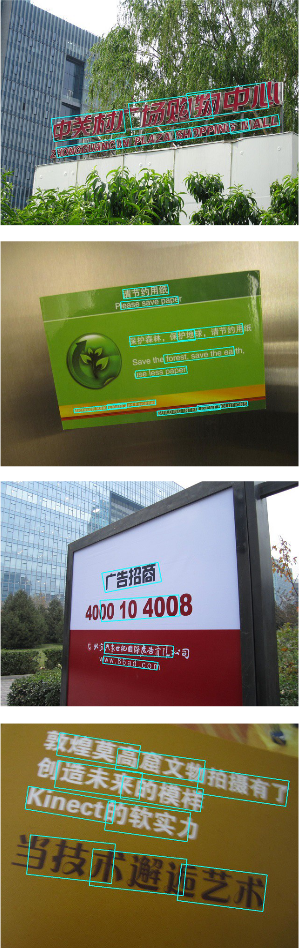}\\
		\end{minipage}%
	}%
	\subfigure[GFNet]{
		\begin{minipage}[t]{0.2\linewidth}
			\centering
			\includegraphics[width=1.02\linewidth]{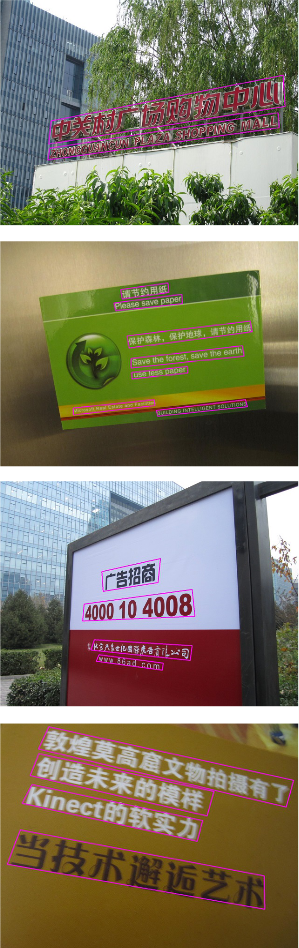}\\
		\end{minipage}%
	}%
	\centering
	\caption{Representative detection results of long texts on MSRA-TD500. (a) Dense boxes generated by EAST, (b) visual detection  results of Standard NMS, (c) visual detection  results of Cascaded NMS, (d) visual detection  results of Skew NMS, (e) visual detection  results of Locality-Aware NMS, (f) visual detection results of our method (GFNet).}
	\label{fig:result_1}
\end{figure*}

\textbf{MSRA-TD500.} This dataset contains both English and Chinese texts, which are all text-level annotated. During testing, we limit the long side of input images for the detector EAST to 1024. The text scale in this dataset varies significantly, even existing many long texts. Therefore, regression-based text detection methods, such as EAST~\citep{EAST}, usually perform poorly on this dataset. For line-level long text detection in MSRA-TD500, the dense boxes generated by EAST ~\citep{EAST} are generally inaccurate, and the prediction errors are large. Hence, the original EAST reported in  ~\citep{EAST} only achieved 76.08\%  of F-measure, even though the improved EAST with Locality-Aware NMS only achieved 80.21\% of F-measure. Locality-Aware NMS based on duplicate fusion has more advantages than traditional NMS algorithms based on duplicate removal for processing the inaccurate detection boxes. As listed in Tab.~\ref{table:TD500}, Locality-Aware NMS outperforms Polygon NMS by 1.04\%  in terms of F-measure. By learning to fuse dense boxes, our GFNet greatly improves the final detection results. Specifically, our GFNet respectively outperforms EAST*(Locality-Aware NMS) by 3.91\% 
and outperforms EAST*(Polygon NMS) by 4.95\% in terms of F-measure.  Equipped with the proposed GFNet, \textbf{EAST*} can achieve a promising F-measure score 84.12\%, surpassing the state-of-the-art methods, such as ATRR~\citep{CVPR19_ATRR}, CRAFT~\citep{CRAFT}, and PAN~\citep{PSENet_v2}.


\begin{table}[htbp]
	\begin{center}
		\renewcommand{\arraystretch}{1.0}
		\caption{Experimental results on ICDAR17-MLT. The symbol $^*$  means that the model is re-implemented. The “R”, “P” and “F” represent the Recall, Precision and F-measure, respectively. “P-NMS”, “S-NMS” and “L-NMS” represent the Polygon NMS, Skew NMS and Locality-Aware NMS, respectively.}
		\label{table:MLT}
		\resizebox{0.485\textwidth}{!}{
			\begin{tabular}{|l||c|c|c|}
				\hline
				\textbf{Methods}& \textbf{R}& \textbf{P} & \textbf{F}\\
				\hline
				Ma et al. \citep{RRPN} &55.50 &71.17 &62.37\\ 
				He et al. \citep{MOML} &57.9 &76.7  &66.0\\ 
				Border \citep{ASTD} &60.6 &73.9  &66.6\\ 
				Corner.\citep{corner} &55.6 &\textbf{83.8} &66.8\\
				FOTS\citep{FOTS} &\textbf{61.04} &80.95 &67.25\\
				DRRG~\citep{DRRG}&84.69 &74.99 &67.31\\
				LOMO\citep{CVPR19_LOMO} &60.6 &78.8 &68.5\\
				MCN-4\citep{MCN_ijcv} &859.54 &81.69 & 68.88\\
				\hline
				\hline
				\textbf{EAST$ ^{*} $}(P-NMS)&56.23&76.04&64.65\\
				\textbf{EAST$ ^{*} $}(S-NMS)&52.21&70.45&59.97\\
				\textbf{EAST$ ^{*} $}(L-NMS)&58.03&77.95&66.53 \\
				\textbf{EAST$ ^{*} $(GFNet)}&60.04&80.13&\textbf{68.65}\\
				\hline
		\end{tabular}}
	\end{center}%
\end{table}

\medskip

\textbf{ICDAR2017-MLT.} This dataset is a large-scale multi-lingual text dataset, which contains texts of various sizes and orientations. It contains a considerable number of small texts and a large number of long texts. During testing, we limit the long side of input images to 1,920. The detailed results are listed in Tab.~\ref{table:MLT}. From Tab.~\ref{table:MLT}, we can see that Skew NMS still has the worst performance in terms of F-measure. On this dataset, the weighted fusion strategy~(Locality-Aware NMS) is still more effective than duplicate removal strategies~(\eg, Polygon NMS, Skew NMS). As listed in Tab.~\ref{table:MLT}, Locality-Aware NMS outperforms Polygon NMS by 1.88\% in terms of F-measure. On this complex scene text dataset, our GFNet demonstrates superior effectiveness against traditional NMS algorithms (including Polygon NMS, Skew NMS,  Locality-Aware NMS). From Tab.~\ref{table:MLT}, we can see that our GFNet respectively outperforms Locality-Aware NMS  by 2.21\% and Polygon NMS by 4.0\% in terms of F-measure. In addition, Our EAST*(GFNet) achieves a promising  F-measure score 84.12\%, 
surpassing many state-of-the-art methods, such as LOMO~\citep{CVPR19_LOMO}, ATRR ~\citep{CVPR19_ATRR}, CRAFT~\citep{CRAFT}  and DRRG ~\citep{DRRG}.

\begin{table}[htbp]
	\begin{center}
		\renewcommand{\arraystretch}{1.0}
		\caption{Time consumption (ms) comparison of several NMS algorithms. “P-NMS”, “S-NMS” and “L-NMS” represent the Polygon NMS, Skew NMS and Locality-Aware NMS, respectively.}
		\label{table:speed}
		\resizebox{0.49\textwidth}{!}{
			\begin{tabular}{|l||c|c|c|}
				\hline
				\textbf{Dataset}& \textbf{ICDAR2015}& \textbf{MSRA-TD500} & \textbf{ICDAR2017-MLT}\\
				\hline
				P-NMS&223&659&1578\\
				\hline 
				S-NMS&15.7&23.0&38.0\\
				\hline       
				L-NMS&117&393 &523\\
				\hline 
				\textbf{GFNet}&143&557 &645\\
				\hline
		\end{tabular}}
	\end{center}%
\end{table}

\begin{figure}[htbp]
	\begin{center}
		\includegraphics[width=0.98\linewidth]{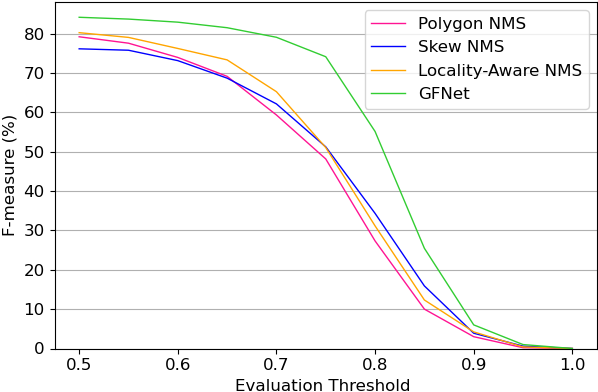}
		\caption{F-measure value of different evaluation thresholds. In our experiment, we sample evaluation threshold with a step of 0.05.} 
		\label{fig:carve_line}
	\end{center}
\vspace{-1.2em}
\end{figure}

\textbf{Speed Analysis.} we adopt the traditional NMS algorithms (including Polygon NMS, Skew NMS, and Locality-Aware NMS) and our GFNet to deal with the dense boxes detected by text detector EAST, respectively. For a fair comparison, all algorithms are implemented in Python. The efficiency of different NMS is compared and analyzed, as listed in Tab. 4. According to Tab.~\ref{table:speed}, Polygon NMS is extremely expensive and time-consuming among these NMS algorithms because it has O(n$ ^{2} $) computational complexity. In contrast, Skew NMS achieves a fantastic speed yet significantly degrades its performance as listed in Tab.~\ref{table:ICDAR15}, Tab.~\ref{table:TD500} and Tab.~\ref{table:MLT}. Although our GFNet is not the fastest, the improvement of detection performance is promising, as listed in Tab.~\ref{table:ICDAR15}, Tab.~\ref{table:TD500} and Tab.~\ref{table:MLT}. Considering the promising improvement of performance, a slight increase in time consumption is negligible.

\begin{table*}[htbp]
	\begin{center}
		\caption{ Experimental results of OBB and HBB task on DOTA datasets. OBB is the task-oriented leaderboard, and HBB is the task-horizontal leaderboard. The short names for categories are defined as: PL-Plane, BD-Baseball diamond, BR-Bridge, GTF-Ground field track, SV-Small vehicle, LV-Large vehicle, SH-Ship, TC-Tennis court, BC-Basketball court, ST-Storage tank, SBF-Soccer-ball field, RA-Roundabout, HA-Harbor, SP-Swimming pool, and HC-Helicopter. The symbol $^*$  means that the model is re-implemented.}
		\label{table:DOTA} 
		\resizebox{1.0\textwidth}{!}{
			\begin{tabular}{|c|c|c|c|c|c|c|c|c|c|c|c|c|c|c|c|c|}
				\hline
				Methods&mAP&PL&BD&BR&GTF&SV&LV&SH&TC&BC&ST&SBF&RA&HA&SP&HC\\
				\hline
				\multicolumn{17}{|c|}{\textbf{OBB}} \\
				\hline
				SSD \citep{RRPN} &10.59&39.83&9.09&0.64&13.18&0.26&0.39&1.11&16.24&27.57&9.23&27.16&9.09&3.03&1.05&1.01\\
				YOLOv2 \citep{YOLO2}&21.39&39.57&20.29&36.58&23.42&8.85&2.09&4.82&44.34&38.35&34.65&16.02&37.62&47.23&25.5&7.45\\
				R-FCN~\citep{R-FCN}&26.79&37.8&38.21&3.64&37.26&6.74&2.6&5.59&22.85&46.93&66.04&33.37&47.15&10.6&25.19&17.96\\ 
				FR-H~\citep{Faster-rcnn}&36.29&47.16&61&9.8&51.74&14.87&12.8&6.88&56.26&59.97&57.32&47.83&48.7&8.23&37.25&23.05\\
				FR-O~\citep{FR-O}&52.93&79.09&69.12&17.17&63.49&34.2&37.16&36.2&89.19&69.6&58.96&49.4&52.52&46.69&44.8&46.3\\
				R-DFPN \citep{R-DFPN}  & 57.94& 80.92 & 65.82 & 33.77 & 58.94 & 55.77 & 50.94 & 54.78 & 90.33 & 66.34 & 68.66 & 48.73 & 51.76 & 55.10 & 51.32 & 35.88 \\
				R$ ^{2} $CNN~\citep{R2CNN}&60.67&80.94&65.75&35.34&67.44&59.92&50.91&55.81&90.67&66.92&72.39&55.06&52.23&55.14&53.35&48.22\\
				RRPN~\citep{RRPN}&61.01&88.52&71.20&31.66&59.30&51.85&56.19&57.25&\textbf{90.81}&72.84&67.38&56.69&52.84&53.08&51.94&53.58\\
				ICN~\citep{ICN}&68.20&81.40&74.30&47.70&70.30&64.90&67.80&70.00&90.80&79.10&78.20&53.60&62.90&67.00&64.20&50.20\\
				RoI-Transformer~\citep{RoI_Transformer} &69.56&88.64 &78.52 &43.44 &75.92 &68.81 &\textbf{73.68} &83.59 &90.74 &77.27 &81.46 &58.39 &53.54 &62.83 &58.93 &47.67\\
				DRN~\citep{DRN}&70.70&88.91 &80.222 &43.52 &63.35 &\textbf{73.48} &70.69 &\textbf{84.94} &90.14 &\textbf{83.85} &84.11 &50.12 &58.41 &67.62 &68.60 &52.50\\
				\hline
				R$ ^{2} $CNN++$ ^{*} $(R-NMS)& 69.31 &89.72&80.62&45.31&72.18&66.26&56.77&63.99&90.56&77.80&85.50&60.91&61.37&65.44&68.23 &54.89\\
				R$ ^{2} $CNN++$ ^{*} $(GFNet)&\textbf{71.58}&\textbf{90.25}&\textbf{83.31}&\textbf{51.94}&\textbf{77.12}&65.47&58.21&61.53&90.65&82.14&\textbf{86.12}&\textbf{65.33}&\textbf{63.91}&\textbf{70.56}&\textbf{69.49}&\textbf{57.69}\\
				\hline
				\hline
				\multicolumn{17}{|c|}{\textbf{HBB}} \\
				\hline
				SSD \citep{SSD} & 10.94 & 44.74 & 11.21 & 6.22 & 6.91 & 2.00 & 10.24 & 11.34 & 15.59 & 12.56 & 17.94 & 14.73 & 4.55 & 4.55 & 0.53 & 1.01 \\
				YOLOv2 \citep{YOLO2} & 39.20 & 76.90 & 33.87 & 22.73 & 34.88 & 38.73 & 32.02 & 52.37 & 61.65 & 48.54 & 33.91 & 29.27 & 36.83 & 36.44 & 38.26 & 11.61 \\
				R-FCN~\citep{R-FCN} & 47.24 & 79.33 & 44.26 & 36.58 & 53.53 & 39.38 & 34.15 & 47.29 & 45.66 & 47.74 & 65.84 & 37.92 & 44.23 & 47.23 & 50.64 & 34.90 \\
				FR-H \citep{Faster-rcnn}& 60.46 & 80.32 & 77.55 & 32.86 & 68.13 & 53.66 & 52.49 & 50.04 & 90.41 & 75.05 & 59.59 & 57.00 & 49.81 & 61.69 & 56.46 & 41.85 \\
				FPN \citep{FPN}  & 72.00 & 88.70 & 75.10 & 52.60 & 59.20 & 69.40 & {\bf 78.80} & {\bf 84.50} & 90.60 & 81.30 & 82.60 & 52.50 & 62.10 & 76.60 & 66.30 & 60.10\\
				ICN \citep{ICN} & 72.50 & 90.00 & 77.70 & 53.40 & {\bf 73.30} & {\bf 73.50} & 65.00 & 78.20 & \textbf{90.80} & 79.10 & 84.80 & 57.20 & 62.10 & 73.50 & 70.20 & 58.10 \\
				\hline
				R$ ^{2} $CNN++$ ^{*} $(Standard NMS)& 74.32&90.05&80.99&55.68&70.58&71.95&75.11&78.13&90.70&80.15&85.79&59.25&62.25&75.34&78.47&60.28\\
				R$ ^{2} $CNN++$ ^{*} $(Soft NMS G)& 75.03&89.41&82.25&56.75&71.82&72.46&74.97&78.47&90.64&81.37&85.88&62.73&63.14&\textbf{76.86}&78.24&61.22\\
				R$ ^{2} $CNN++$ ^{*} $(Soft NMS L)& 75.15&89.57&82.43&56.54&71.69&72.45&74.83&78.54&90.67&82.23&85.06&63.41&\textbf{63.23}&76.71&78.43&61.44\\
				R$ ^{2} $CNN++$ ^{*} $(GFNet)&\textbf{75.87}&\textbf{90.42}&\textbf{85.25}&\textbf{60.96}&72.96&70.89&73.83&76.55&90.79&\textbf{86.37}&\textbf{86.66}&\textbf{64.84}&61.87&73.75&\textbf{79.41}&\textbf{63.43}\\
				\hline
		\end{tabular}}
	\end{center}%
\end{table*}

\textbf{Sensitivity Analysis for Evaluation Threshold.}
We conduct sensitivity analysis on the line-level text detection dataset MSRA-TD500. By setting different evaluation thresholds (IoU threshold),  we explore the sensitivity of the detection results generated by different NMS algorithms to the evaluation threshold. Generally, the evaluation threshold is set to 0.5 by default on evaluating text detection results. But, it is not a very strict threshold value. Under the threshold of 0.5 in evaluation, even if the prediction of the text detector does not entirely cover the whole text instance, it may be considered correct. Generally speaking, a more stringent threshold can better reflect the accuracy of detection results. Therefore, to further compare the accuracy of the detection results generated by different NMS algorithms, we gradually increase the evaluation threshold with a step of 0.05 to explore the change of F-measure values for different NMS algorithms. As shown in Fig.~\ref{fig:carve_line}, with the increment of evaluation threshold, the performances of traditional NMS algorithms (including Polygon NMS, Skew NMS,  Locality-Aware NMS) all drop rapidly. Compared with the conventional NMS algorithms, the F-measure value of our GFNet does not decrease significantly when the evaluation threshold is less than 0.7.  When the evaluation threshold is set to 0.8, our GFNet still has 55.12\% F-measure, outperforming Locality-Aware NMS by 23.98\% in terms of F-measure. In addition, the curve corresponding to our GFNet is always above the curve of the other three methods (including Polygon NMS, Skew NMS,  Locality-Aware NMS), which fully proves that the final detection boxes of our GFNet are more accurate than the other three NMS algorithms.

\begin{figure*}[t!p]
	\subfigcapskip=0pt
	\centering
	\subfigure[Dense Boxes]{
		\begin{minipage}[b]{0.98\linewidth}
			\includegraphics[width=1\linewidth]{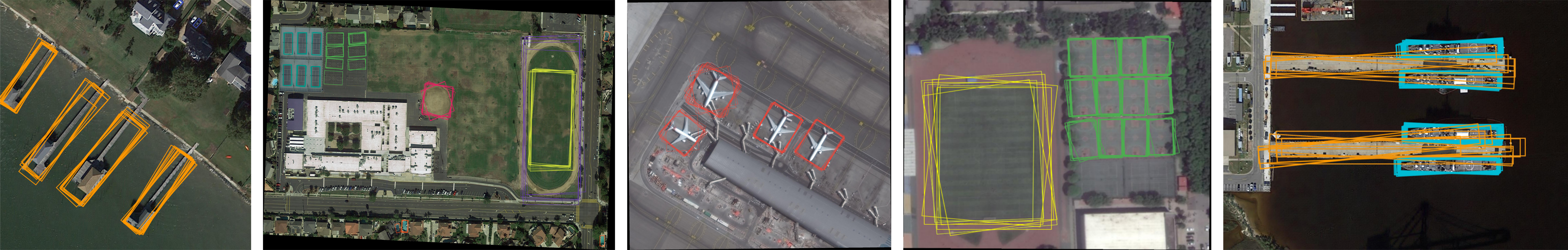}\\
			\vspace{-0.8em}
	\end{minipage}}
	\subfigure[R-NMS]{
		\begin{minipage}[b]{0.98\linewidth}
			\includegraphics[width=1\linewidth]{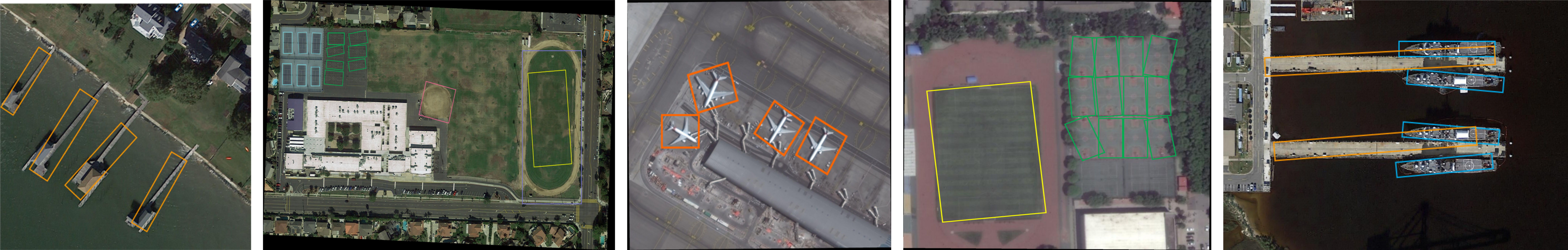}\\
			\vspace{-0.8em}
	\end{minipage}}
	\subfigure[GFNet]{
		\begin{minipage}[b]{0.98\linewidth}
			\includegraphics[width=1\linewidth]{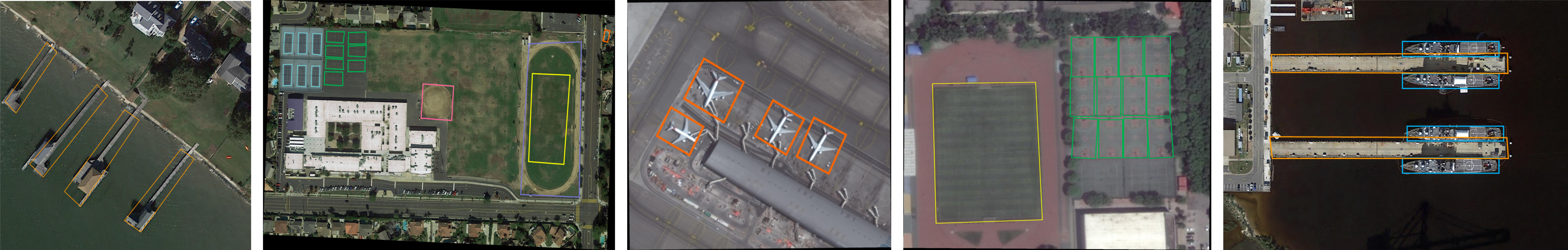}\\
			\vspace{-0.8em}
	\end{minipage}}
	\caption{Representative results on DOTA. The first row is the dense detection boxes generated by the network of R$ ^{2} $CNN++ ; the Second row is the result of rotation non-maximum suppression (R-NMS) for removing duplicate boxes; the Bottom row is the detection results of our GFNet for fusing duplicate boxes. Apparently, our GFNet can significantly improve the performance of objects with large sizes and arbitrary directions. Note, we only visualized some detection boxes of each object with systematic sampling.}
	\label{fig:result_all_2}
	\vspace{-1.2em}
\end{figure*}

\textbf{Visual results.}  Fig.~\ref{fig:result_1} shows some  representative visual results of different NMS algorithms. For detecting long texts, the direct regression-based text detection method (\eg, EAST~\citep{EAST}) usually can't produce an accurate detection box well covering the whole text instance. In this case, traditional NMS algorithms (including Polygon NMS, Skew NMS, Locality-Aware NMS) can not effectively use the relation of these dense detection boxes to further optimize the final detection result. As shown in Fig.~\ref{fig:result_1} (a)-(d), the traditional NMS algorithm fail to deal with these inaccurate detection boxes. In contrast, by learning a graph-based fusion model, our GFNet can automatically fuse detection boxes into accurate instance boxes, as shown in Fig.~\ref{fig:result_1} (d).

\subsection{\normalsize{Object Detection in Aerial Images}}\label{object_detect_exp}
To verify the effectiveness of our GFNet for general multi-oriented object detection, we conduct experiments on the popular aerial image dataset DOTA~\citep{DOTA}. This dataset contains objects with a wide variety of scales, orientations, and shapes. There are two detection tasks for DOTA: horizontal bounding boxes (HBB) and oriented bounding boxes (OBB). We use the Python implementation of R$ ^{2}$CNN++~\citep{R2CNN++} (anchor-based), which uses the rotation non-maximum suppression (R-NMS)~\citep{R2CNN++} by default to remove duplicates. In addition, it can process OBB tasks and HBB tasks at the same time. After we get dense detection boxes generated by R$ ^{2}$CNN++, we adopt different NMS algorithms to remove duplicate boxes.


\textbf{OBB Task.} In this task, we compare our GFNet with R$^{2}$CNN++*(R$ ^{2}$CNN++ equipped with R-NMS). As listed in Tab.~\ref{table:DOTA}, our GFNet outperforms R-NMS by  2.27\% in terms of mAP.  And, GFNet can further improves the performance of detector R $^ {2} $CNN++ on OBB task compared with R-NMS. With GFNet, the performance of R $^ {2} $CNN++* can reach 71.58\% mAP, surpassing state-of-the-art method DRN~\citep{DRN} with a great margin. Also, our method~(R $^ {2} $CNN++(GFNet)*) outperforms RoI-Transformer~\citep{RoI_Transformer} by 2.02\% in terms of mAP, and outperforms RRPN~\citep{RRPN} by 10.57\% in terms of mAP.

\textbf{HBB Task.}  In this task, R$ ^{2}$CNN++ will generate abundant horizontal bounding boxes. We adopt the Standard NMS, Soft NMS, and GFNet to process these horizontal detection boxes in our experiments. The detailed results are listed in Tab.~\ref{table:DOTA}, in which ``Soft NMS G'' denotes Soft NMS with Gaussian weighting and ``Soft NMS L'' denotes Soft NMS with linear weighting. Specifically,  our GFNet processing still tops the rank. Our GFNet outperforms Standard NMS by 1.55\%  and outperforms Soft NMS with linear weighting by 0.72\% in terms of mAP. In addition, our method~(R $^ {2} $CNN++*(GFNet)) outperforms ICN~\citep{ICN} by 3.37\% in terms of mAP, and outperforms YOLOv2~\citep{YOLO2} by 36.67\% in terms of mAP.

\textbf{Visual results.}  Fig.~\ref{fig:result_all_2} demonstrates some representative  detection results generated by R-NMS and our GFNet. For large-scale objects,  dense boxes generated by R$ ^{2}$CNN++ is very inaccurate. In this case, if we only perform duplicate removal as R-NMS, the results are always unsatisfactory, as shown in the second row of Fig.~\ref{fig:result_all_2}. In contrast, our GFNet can learn to adaptively fuse these dense boxes in the same object instance, generating more accurate detection results even if the dense boxes generated by the detector are not accurate, as shown in the third row of Fig.~\ref{fig:result_all_2}.

\subsection{\normalsize{Weakness Analysis}}~\label{Weakness}
As demonstrated in the above experiments, our GFNet is robust to inaccurate dense boxes, which extensively exist in the scenario of large-scale object detection. This is also verified by the state-of-the-art performances of our method on MSRA-TD500 and DOTA (both two datasets consist of abundant large objects in images). For very small objects, false and missing detection are more likely to occur than inaccurate detection boxes. In this case, GFNet may also have poor performance. In addition, if the object instances are too dense and close to each other, our locality-aware clustering algorithm may produce wrong clusters, which will lead to wrong detection results. Therefore, in Tab.~\ref{table:DOTA}, the detection performance of our GFNet on some objects detection is degraded, such as small vehicles and ships in OBB task or harbor and a large vehicle in HBB task.

\section{Conclusions}\label{Conclusion}
In deep object detection, existing NMS methods via simple intersection-over-Union (IoU) metrics tend to underperform on multi-oriented and long size object detection due to the degraded quality of dense detection boxes and not explicit exploration of the context information. This paper presents a graph-based fusion network named GFNet, for multi-oriented object detection. Our GFNet is extensible and adaptively fuse dense detection boxes to detect more accurate and holistic multi-oriented object instances. Extensive experiments on both public available multi-oriented text datasets (including MSRA-TD500, ICDAR2015, ICDAR2017-MLT) and multi-oriented object datasets (DOTA) verify the effectiveness and robustness of our method against general NMS methods. In the future, we are interested in designing an end-to-end trainable detection system based on our GFNet to improve the practicability and efficiency of the method.

\bibliographystyle{sn-basic.bst}
\bibliography{sn-bibliography}

\begin{thebibliography}{76}
\providecommand{\natexlab}[1]{#1}
\providecommand{\url}[1]{{#1}}
\providecommand{\urlprefix}{URL }
\providecommand{\doi}[1]{\url{https://doi.org/#1}}
\providecommand{\eprint}[2][]{\url{#2}}
 \bibcommenthead

\bibitem[{Azimi et~al(2018)Azimi, Vig, Bahmanyar, K{\"{o}}rner, and
  Reinartz}]{ICN}
Azimi SM, Vig E, Bahmanyar R, et~al (2018) Towards multi-class object detection
  in unconstrained remote sensing imagery. In: Jawahar CV, Li H, Mori G, et~al
  (eds) {ACCV}, pp 150--165

\bibitem[{Baek et~al(2019)Baek, Lee, Han, Yun, and Lee}]{CRAFT}
Baek Y, Lee B, Han D, et~al (2019) Character region awareness for text
  detection. In: CVPR, pp 9365--9374

\bibitem[{Bodla et~al(2017)Bodla, Singh, Chellappa, and Davis}]{soft_nms}
Bodla N, Singh B, Chellappa R, et~al (2017) Soft-nms - improving object
  detection with one line of code. In: {ICCV}, pp 5562--5570

\bibitem[{Bruna et~al(2014)Bruna, Zaremba, Szlam, and LeCun}]{GCN_1}
Bruna J, Zaremba W, Szlam A, et~al (2014) Spectral networks and locally
  connected networks on graphs. In: {ICLR}

\bibitem[{Chen et~al(2021)Chen, Yang, Zhang, Zhang, and Sun}]{PAQ}
Chen L, Yang T, Zhang X, et~al (2021) Points as queries: Weakly semi-supervised
  object detection by points. In: CVPR, pp 8823--8832

\bibitem[{Dai et~al(2016)Dai, Li, He, and Sun}]{R-FCN}
Dai J, Li Y, He K, et~al (2016) {R-FCN:} object detection via region-based
  fully convolutional networks. In: NIPS, pp 379--387

\bibitem[{Dai et~al(2018)Dai, Huang, Gao, Xu, Chen, Guo, and Qiu}]{FTSN}
Dai Y, Huang Z, Gao Y, et~al (2018) Fused text segmentation networks for
  multi-oriented scene text detection. In: ICPR, pp 3604--3609

\bibitem[{Defferrard et~al(2016)Defferrard, Bresson, and Vandergheynst}]{GCN_2}
Defferrard M, Bresson X, Vandergheynst P (2016) Convolutional neural networks
  on graphs with fast localized spectral filtering. In: NIPS, pp 3837--3845

\bibitem[{Deng et~al(2018)Deng, Liu, Li, and Cai}]{PixelLink}
Deng D, Liu H, Li X, et~al (2018) {PixelLink}: Detecting scene text via
  instance segmentation. In: AAAI, pp 6773--6780

\bibitem[{Ding et~al(2018)Ding, Xue, Long, Xia, and Lu}]{RoI_Transformer}
Ding J, Xue N, Long Y, et~al (2018) Learning roi transformer for detecting
  oriented objects in aerial images. CoRR abs/1812.00155

\bibitem[{Duan et~al(2019)Duan, Bai, Xie, Qi, Huang, and Tian}]{CenterNet}
Duan K, Bai S, Xie L, et~al (2019) Centernet: Keypoint triplets for object
  detection. In: {ICCV}. {IEEE}, pp 6568--6577

\bibitem[{Felzenszwalb et~al(2010)Felzenszwalb, Girshick, McAllester, and
  Ramanan}]{NMS}
Felzenszwalb PF, Girshick RB, McAllester DA, et~al (2010) Object detection with
  discriminatively trained part-based models. {IEEE} Trans Pattern Anal Mach
  Intell 32(9):1627--1645

\bibitem[{Girshick(2015)}]{Fast_RCNN}
Girshick RB (2015) Fast {R-CNN}. In: {ICCV}. {IEEE} Computer Society, pp
  1440--1448

\bibitem[{He et~al(2016)He, Zhang, Ren, and Sun}]{ResNet}
He K, Zhang X, Ren S, et~al (2016) Deep residual learning for image
  recognition. In: {CVPR}, pp 770--778

\bibitem[{He et~al(2021)He, Liao, Yang, Zhong, Tang, Cheng, Yao, Wang, and
  Bai}]{MOST}
He M, Liao M, Yang Z, et~al (2021) {MOST:} {A} multi-oriented scene text
  detector with localization refinement. In: CVPR, pp 8813--8822

\bibitem[{He et~al(2017)He, Zhang, Yin, and Liu}]{DDR}
He W, Zhang XY, Yin F, et~al (2017) Deep direct regression for multi-oriented
  scene text detection. In: ICCV, pp 745--753

\bibitem[{He et~al(2018)He, Zhang, Yin, and Liu}]{MOML}
He W, Zhang X, Yin F, et~al (2018) Multi-oriented and multi-lingual scene text
  detection with direct regression. {IEEE} Trans Image Processing
  27(11):5406--5419

\bibitem[{He et~al(2019)He, Zhu, Wang, Savvides, and Zhang}]{Softer-NMS}
He Y, Zhu C, Wang J, et~al (2019) Bounding box regression with uncertainty for
  accurate object detection. In: {CVPR}, pp 2888--2897

\bibitem[{Hosang et~al(2017)Hosang, Benenson, and Schiele}]{learn_nms}
Hosang JH, Benenson R, Schiele B (2017) Learning non-maximum suppression. In:
  {CVPR}, pp 6469--6477

\bibitem[{Hou et~al(2020)Hou, Zhu, Liu, Sheng, Wu, Wang, and Yin}]{HAM}
Hou JB, Zhu X, Liu C, et~al (2020) {HAM:} hidden anchor mechanism for scene
  text detection. {IEEE} Trans Image Process 29:7904--7916

\bibitem[{Hu et~al(2018)Hu, Gu, Zhang, Dai, and Wei}]{Relation_Networks}
Hu H, Gu J, Zhang Z, et~al (2018) Relation networks for object detection. In:
  {CVPR}, pp 3588--3597

\bibitem[{Huang et~al(2015)Huang, Yang, Deng, and Yu}]{DenseBox}
Huang L, Yang Y, Deng Y, et~al (2015) Densebox: Unifying landmark localization
  with end to end object detection. arXiv preprint arXiv:150904874

\bibitem[{Jiang et~al(2017)Jiang, Zhu, Wang, Yang, Li, Wang, Fu, and
  Luo}]{R2CNN}
Jiang Y, Zhu X, Wang X, et~al (2017) {R2CNN:} rotational region {CNN} for
  orientation robust scene text detection. CoRR abs/1706.09579

\bibitem[{Kang et~al(2017)Kang, Kim, and Yoo}]{KangKY17}
Kang C, Kim G, Yoo SI (2017) Detection and recognition of text embedded in
  online images via neural context models. In: {AAAI}, pp 4103--4110

\bibitem[{Karatzas et~al(2015)Karatzas, Gomez{-}Bigorda, Nicolaou, Ghosh, and
  et~al.}]{IC15}
Karatzas D, Gomez{-}Bigorda L, Nicolaou A, et~al (2015) {ICDAR} 2015
  competition on robust reading. In: {ICDAR}, pp 1156--1160

\bibitem[{Kingma and Ba(2014)}]{ADAM}
Kingma DP, Ba J (2014) Adam: {A} method for stochastic optimization. arXiv
  preprint arXiv:14126980

\bibitem[{Kipf and Welling(2017)}]{GCN}
Kipf TN, Welling M (2017) Semi-supervised classification with graph
  convolutional networks. In: {ICLR}

\bibitem[{Liao et~al(2018{\natexlab{a}})Liao, Shi, and Bai}]{textboxes++}
Liao M, Shi B, Bai X (2018{\natexlab{a}}) Textboxes++: A single-shot oriented
  scene text detector. IEEE Transactions on Image Processing 27(8):3676--3690

\bibitem[{Liao et~al(2018{\natexlab{b}})Liao, Zhu, Shi, Xia, and Bai}]{RRD}
Liao M, Zhu Z, Shi B, et~al (2018{\natexlab{b}}) Rotation-sensitive regression
  for oriented scene text detection. In: {CVPR}, pp 5909--5918

\bibitem[{Lin et~al(2017)Lin, Doll{\'{a}}r, Girshick, He, Hariharan, and
  Belongie}]{FPN}
Lin T, Doll{\'{a}}r P, Girshick RB, et~al (2017) Feature pyramid networks for
  object detection. In: CVPR, pp 936--944

\bibitem[{Liu et~al(2016)Liu, Anguelov, Erhan, Szegedy, Reed, Fu, and
  Berg}]{SSD}
Liu W, Anguelov D, Erhan D, et~al (2016) {SSD:} {Single} shot multibox
  detector. In: ECCV, pp 21--37

\bibitem[{Liu et~al(2018{\natexlab{a}})Liu, Liang, Yan, Chen, Qiao, and
  Yan}]{FOTS}
Liu X, Liang D, Yan S, et~al (2018{\natexlab{a}}) {FOTS: Fast} oriented text
  spotting with a unified network. In: {CVPR}, pp 5676--5685

\bibitem[{Liu et~al(2017)Liu, Jin, Zhang, and Zhang}]{CTW1500}
Liu Y, Jin L, Zhang S, et~al (2017) Detecting curve text in the wild: New
  dataset and new solution. CoRR abs/1712.02170

\bibitem[{Liu et~al(2018{\natexlab{b}})Liu, Lin, S.Yang, Feng, Lin, and
  Ling~Goh}]{MCN}
Liu Z, Lin G, S.Yang, et~al (2018{\natexlab{b}}) Learning markov clustering
  networks for scene text detection. In: CVPR, pp 6936--6944

\bibitem[{Liu et~al(2020)Liu, Lin, and Goh}]{MCN_ijcv}
Liu Z, Lin G, Goh WL (2020) Bottom-up scene text detection with markov
  clustering networks. Int J Comput Vis 128(6):1786--1809

\bibitem[{Long et~al(2018)Long, Ruan, Zhang, He, Wu, and Yao}]{TextSnake}
Long S, Ruan J, Zhang W, et~al (2018) Textsnake: {A} flexible representation
  for detecting text of arbitrary shapes. In: {ECCV}, pp 19--35

\bibitem[{Lyu et~al(2018)Lyu, Yao, Wu, Yan, and Bai}]{corner}
Lyu P, Yao C, Wu W, et~al (2018) Multi-oriented scene text detection via corner
  localization and region segmentation. In: CVPR, pp 7553--7563

\bibitem[{Ma et~al(2018)Ma, Shao, Ye, Wang, Wang, Zheng, and Xue}]{RRPN}
Ma J, Shao W, Ye H, et~al (2018) Arbitrary-oriented scene text detection via
  rotation proposals. {IEEE} Trans Multimedia 20(11):3111--3122

\bibitem[{Nayef et~al(2017)Nayef, Yin, Bizid, Choi, Feng, Karatzas, and
  et~al.}]{MLT}
Nayef N, Yin F, Bizid I, et~al (2017) {ICDAR2017} robust reading challenge on
  multi-lingual scene text detection and script identification - {RRC-MLT}. In:
  {ICDAR}, pp 1454--1459

\bibitem[{Pan et~al(2020)Pan, Ren, Sheng, Dong, Yuan, Guo, Ma, and Xu}]{DRN}
Pan X, Ren Y, Sheng K, et~al (2020) Dynamic refinement network for oriented and
  densely packed object detection. In: {CVPR}, pp 11,204--11,213

\bibitem[{Peng et~al(2022)Peng, Wang, Yue, and Zhang}]{Context-aware}
Peng J, Wang H, Yue S, et~al (2022) Context-aware co-supervision for accurate
  object detection. Pattern Recognit 121:108,199

\bibitem[{Philbin et~al(2007)Philbin, Chum, Isard, Sivic, and
  Zisserman}]{Object_matching}
Philbin J, Chum O, Isard M, et~al (2007) Object retrieval with large
  vocabularies and fast spatial matching. In: {CVPR}

\bibitem[{Qi et~al(2018)Qi, Liu, Shi, and Jia}]{RNN_nms}
Qi L, Liu S, Shi J, et~al (2018) Sequential context encoding for duplicate
  removal. In: NeurIPS, pp 2053--2062

\bibitem[{Redmon et~al(2016)Redmon, Divvala, Girshick, and Farhadi}]{YOLO1}
Redmon J, Divvala SK, Girshick RB, et~al (2016) You only look once: Unified,
  real-time object detection. In: {CVPR}, pp 779--788

\bibitem[{Redmon et~al(2017)Redmon, Farhad, and et~al}]{YOLO2}
Redmon J, Farhad A, et~al (2017) {YOLO9000:} {Better}, faster, stronger. In:
  {CVPR}, pp 6517--6525

\bibitem[{Ren et~al(2017)Ren, He, Girshick, and Sun}]{Faster-rcnn}
Ren S, He K, Girshick RB, et~al (2017) Faster {R-CNN:} {Towards} real-time
  object detection with region proposal networks. {IEEE} Trans Pattern Anal
  Mach Intell 39(6):1137--1149

\bibitem[{Rong et~al(2016)Rong, Yi, and Tian}]{drive1}
Rong X, Yi C, Tian Y (2016) Recognizing text-based traffic guide panels with
  cascaded localization network. In: {ECCV} Workshops, pp 109--121

\bibitem[{Shi et~al(2017)Shi, Bai, and Belongie}]{SegLink}
Shi B, Bai X, Belongie SJ (2017) Detecting oriented text in natural images by
  linking segments. In: CVPR, pp 3482--3490

\bibitem[{Shrivastava et~al(2016)Shrivastava, Gupta, and Girshick}]{OHEM}
Shrivastava A, Gupta A, Girshick RB (2016) Training region-based object
  detectors with online hard example mining. In: {CVPR}, pp 761--769

\bibitem[{Sivic and Zisserman(2003)}]{Video_Google}
Sivic J, Zisserman A (2003) Video google: {A} text retrieval approach to object
  matching in videos. In: {ICCV}, pp 1470--1477

\bibitem[{Tian et~al(2019)Tian, Shu, Lyu, Li, Zhou, Shen, and Jia}]{CVPR19_LSA}
Tian Z, Shu M, Lyu P, et~al (2019) Learning shape-aware embedding for scene
  text detection. In: CVPR, pp 4234--4243

\bibitem[{Velickovic et~al(2018)Velickovic, Cucurull, Casanova, Romero,
  Li{\`{o}}, and Bengio}]{GAT}
Velickovic P, Cucurull G, Casanova A, et~al (2018) Graph attention networks.
  In: ICLR

\bibitem[{Wang et~al(2018)Wang, Zhao, Li, Wang, and Tao}]{ITN}
Wang F, Zhao L, Li X, et~al (2018) Geometry-aware scene text detection with
  instance transformation network. In: CVPR

\bibitem[{Wang et~al(2019{\natexlab{a}})Wang, Xie, Li, Hou, Lu, Yu, and
  Shao}]{CVPR19_PSENet}
Wang W, Xie E, Li X, et~al (2019{\natexlab{a}}) Shape robust text detection
  with progressive scale expansion network. In: CVPR, pp 9336--9345

\bibitem[{Wang et~al(2019{\natexlab{b}})Wang, Xie, Song, Zang, Wang, Lu, Yu,
  and Shen}]{PSENet_v2}
Wang W, Xie E, Song X, et~al (2019{\natexlab{b}}) Efficient and accurate
  arbitrary-shaped text detection with pixel aggregation network. In: ICCV, pp
  8439--8448

\bibitem[{Wang et~al(2019{\natexlab{c}})Wang, Jiang, Luo, Liu, Choi, and
  Kim}]{CVPR19_ATRR}
Wang X, Jiang Y, Luo Z, et~al (2019{\natexlab{c}}) Arbitrary shape scene text
  detection with adaptive text region representation. In: CVPR, pp 6449--6458

\bibitem[{Wang et~al(2020)Wang, Xie, Zha, Xing, Fu, and Zhang}]{ContourNet}
Wang Y, Xie H, Zha Z, et~al (2020) Contournet: Taking a further step toward
  accurate arbitrary-shaped scene text detection. In: {CVPR}, pp 11,750--11,759

\bibitem[{Wang et~al(2019{\natexlab{d}})Wang, Zheng, Li, and
  Wang}]{CVPR19_Linkage}
Wang Z, Zheng L, Li Y, et~al (2019{\natexlab{d}}) Linkage based face clustering
  via graph convolution network. In: CVPR, pp 1117--1125

\bibitem[{Xia et~al(2018{\natexlab{a}})Xia, Bai, Ding, Zhu, Belongie, Luo,
  Datcu, Pelillo, and Zhang}]{DOTA}
Xia G, Bai X, Ding J, et~al (2018{\natexlab{a}}) {DOTA:} {A} large-scale
  dataset for object detection in aerial images. In: {CVPR}, pp 3974--3983

\bibitem[{Xia et~al(2018{\natexlab{b}})Xia, Bai, Ding, Zhu, Belongie, Luo,
  Datcu, Pelillo, and Zhang}]{FR-O}
Xia G, Bai X, Ding J, et~al (2018{\natexlab{b}}) {DOTA:} {A} large-scale
  dataset for object detection in aerial images. In: {CVPR}, pp 3974--3983

\bibitem[{Xu et~al(2019)Xu, Wang, Zhou, Wang, Yang, and Bai}]{TextField}
Xu Y, Wang Y, Zhou W, et~al (2019) Textfield: Learning a deep direction field
  for irregular scene text detection. {IEEE} Trans Image Processing
  28(11):5566--5579

\bibitem[{Xue et~al(2018)Xue, Lu, and Zhan}]{ASTD}
Xue C, Lu S, Zhan F (2018) Accurate scene text detection through border
  semantics awareness and bootstrapping. In: ECCV, pp 370--387

\bibitem[{Xue et~al(2019)Xue, Lu, and Zhang}]{MSR}
Xue C, Lu S, Zhang W (2019) {MSR:} multi-scale shape regression for scene text
  detection. In: IJCAI, pp 989--995

\bibitem[{Yang et~al(2018{\natexlab{a}})Yang, Cheng, Zhou, Chen, Qiu, and
  Lin}]{IncepText}
Yang Q, Cheng M, Zhou W, et~al (2018{\natexlab{a}}) Inceptext: A new
  inception-text module with deformable psroi pooling for multi-oriented scene
  text detection. In: {IJCAI}, pp 1071--1077

\bibitem[{Yang et~al(2018{\natexlab{b}})Yang, Sun, Fu, Yang, Sun, Yan, and
  Guo}]{R-DFPN}
Yang X, Sun H, Fu K, et~al (2018{\natexlab{b}}) Automatic ship detection in
  remote sensing images from google earth of complex scenes based on multiscale
  rotation dense feature pyramid networks. Remote Sens 10(1):132

\bibitem[{Yang et~al(2019)Yang, Yang, Yan, Zhang, Zhang, Guo, Sun, and
  Fu}]{R2CNN++}
Yang X, Yang J, Yan J, et~al (2019) Scrdet: Towards more robust detection for
  small, cluttered and rotated objects. In: {ICCV}, pp 8231--8240

\bibitem[{Yao et~al(2012)Yao, Bai, Liu, Ma, and Tu}]{MSRATD500}
Yao C, Bai X, Liu W, et~al (2012) Detecting texts of arbitrary orientations in
  natural images. In: {CVPR}, pp 1083--1090

\bibitem[{Yin et~al(2016)Yin, Zuo, Tian, and Liu}]{Yin-Z}
Yin X.C., Zuo Z, Tian S, et~al (2016) Text detection, tracking and recognition in
  video: {A} comprehensive survey. {IEEE} Trans Image Processing
  25(6):2752--2773

\bibitem[{Zhang et~al(2019)Zhang, Liang, Huang, En, Han, Ding, and
  Ding}]{CVPR19_LOMO}
Zhang C, Liang B, Huang Z, et~al (2019) Look more than once: An accurate
  detector for text of arbitrary shapes. In: CVPR, pp 10,552--10,561

\bibitem[{Zhang et~al(2020)Zhang, Zhu, Hou, Liu, Yang, Wang, and Yin}]{DRRG}
Zhang S.X., Zhu X, Hou J, et~al (2020) Deep relational reasoning graph network for
  arbitrary shape text detection. In: {CVPR}, pp 9696--9705

\bibitem[{Zhang et~al(2021)Zhang, Zhu, Yang, Wang, and Yin}]{TextBPN}
Zhang S.X., Zhu X, Yang C, et~al (2021) Adaptive boundary proposal network for
  arbitrary shape text detection. In: ICCV, pp 1305--1314

\bibitem[{Zhang et~al(2022{\natexlab{a}})Zhang, Zhu, Chen, Hou, and
  Yin}]{Zhang2022ArbitraryST}
Zhang S.X., Zhu X, Chen L, et~al (2022{\natexlab{a}}) Arbitrary shape text
  detection via segmentation with probability maps. IEEE transactions on
  pattern analysis and machine intelligence pp 2736--2750

\bibitem[{Zhang et~al(2022{\natexlab{b}})Zhang, Zhu, Hou, Yang, and
  Yin}]{Zhang2022KernelPN}
Zhang S.X., Zhu X, Hou JB, et~al (2022{\natexlab{b}}) Kernel proposal network for
  arbitrary shape text detection. IEEE transactions on neural networks and
  learning systems pp

\bibitem[{Zhou et~al(2021)Zhou, Yu, Wang, Qian, and Li}]{Instant-Teaching}
Zhou Q, Yu C, Wang Z, et~al (2021) Instant-teaching: An end-to-end
  semi-supervised object detection framework. In: CVPR, pp 4081--4090

\bibitem[{Zhou et~al(2017)Zhou, C.Yao, Wen, Wang, Zhou, He, and Liang}]{EAST}
Zhou X, C.Yao, Wen H, et~al (2017) {EAST:} {An} efficient and accurate scene
  text detector. In: {CVPR}, pp 2642--2651

\bibitem[{Zhu et~al(2018)Zhu, Liao, Yang, and Liu}]{drive2}
Zhu Y, Liao M, Yang M, et~al (2018) Cascaded segmentation-detection networks
  for text-based traffic sign detection. {IEEE} Trans Intell Transp Syst
  19(1):209--219

\end{thebibliography}


\end{document}